\documentclass[11pt]{article}

\usepackage[preprint]{acl}

\usepackage{times}
\usepackage{latexsym}

\usepackage[T1]{fontenc}

\usepackage[utf8]{inputenc}

\usepackage{microtype}

\usepackage{inconsolata}

\usepackage{graphicx}

\usepackage{subcaption}
\usepackage{amsmath} 
\DeclareMathAlphabet{\mathbbold}{U}{bbold}{m}{n}

\usepackage{booktabs}
\usepackage{multirow}
\usepackage{xcolor}
\usepackage{colortbl}
\definecolor{mygray}{gray}{0.90}       
\definecolor{mylightgray}{gray}{0.95}  
\usepackage{arydshln}

\title{Think How to Think: Mitigating Overthinking with Autonomous Difficulty Cognition in Large Reasoning Models}

\author{
 \textbf{Yongjiang Liu\textsuperscript{1}},
 \textbf{Haoxi Li\textsuperscript{1}},
 \textbf{Xiaosong Ma\textsuperscript{2}},
 \textbf{Jie Zhang\textsuperscript{1}},
 \textbf{Song Guo\textsuperscript{1}},
\\
 \textsuperscript{1}The Hong Kong University of Science and Technology,
 \textsuperscript{2}The Hong Kong Polytechnic University
\\
}

\begin{document}
\maketitle
\begin{abstract}
Recent Large Reasoning Models (LRMs) excel at complex reasoning tasks but often suffer from overthinking, generating overly long and redundant reasoning trajectories.
To explore its essence, our empirical analysis reveals that LRMs are primarily limited to recognizing task properties (i.e., difficulty levels) like humans before solving the problem, leading to a one-size-fits-all reasoning strategy. 
This observation motivates a fundamental question: \textit{Can we explicitly bootstrap such ability to alleviate overthinking in LRMs?} To this end, we propose \textbf{T}hink-\textbf{H}ow-to-\textbf{T}hink (\textbf{TH2T}), a novel two-stage fine-tuning strategy that progressively inspires LRMs' difficulty cognition and redundancy cognition of LRMs. 
Specifically, we first inject \textbf{Difficulty Dypnosis} into output prefixes as cues for global, prospective reasoning strategy selection, stimulating the model's sharper sensitivity to task complexity and adaptive control of reasoning depth.
Then, we incorporate \textbf{Redundancy Hypnosis} into in-progress reasoning steps, which serve as local, retrospective signals for behavior correction by identifying and eliminating superfluous reasoning detours. 
Experiments across 7B/14B/32B models demonstrate that \textbf{TH2T} significantly reduces inference costs by over 70\% on easy tasks and 40\% on complex ones without compromising performance. The resultant models exhibit a nascent ability for difficulty-aware reasoning, effectively mitigating behaviors like excessive reflection and looping, thereby paving the way for more cognitively efficient LRMs.
\end{abstract}

\section{Introduction} 
\label{sec:intro}
The emergence of Large Reasoning Models (LRMs) has led to substantial improvements in problem-solving capabilities~\cite{sun2023survey, yu2024natural, team2025kimi}, particularly through the use of elaborate and sequential reasoning enabled by test-time scaling~\cite{snell2025scaling}. Many models, such as OpenAI-O1~\cite{openai2024openaio1card}, Deepseek-R1~\cite{guo2025deepseek}, QwQ~\cite{qwq32b} leverage long Chain-of-Thought~\cite{wei2022chain, wang2022self} to approach complex tasks step-by-step, enabling a human-like thinking pattern with reflection, backtracking, and self-validation~\cite{xie2025logic, liu2025understandingr1zeroliketrainingcritical}. This reasoning paradigm demonstrably enhances the models’ problem-solving capabilities and yields promising results. 
However, lengthy responses introduce the problem of ``overthinking''~\cite{chen2025think23overthinkingo1like}, wherein model performance initially improves with extended reasoning chains but deteriorates beyond a certain length. This inefficiency is starkly evident when trivial queries, such as ``\textit{What is 1 plus 1?}'', trigger disproportionately complex reasoning paths. Such redundant deliberation not only inflates computational cost and latency but also degrades the user experience. These lead us to a central inquiry: \textbf{\textit{Do LRMs possess the cognitive judgment to dynamically regulate their reasoning depth in response to varying problem difficulty?}}

\begin{figure}[t]
\centering
\subfloat[GSM8K]{\includegraphics[width = 0.235\textwidth]{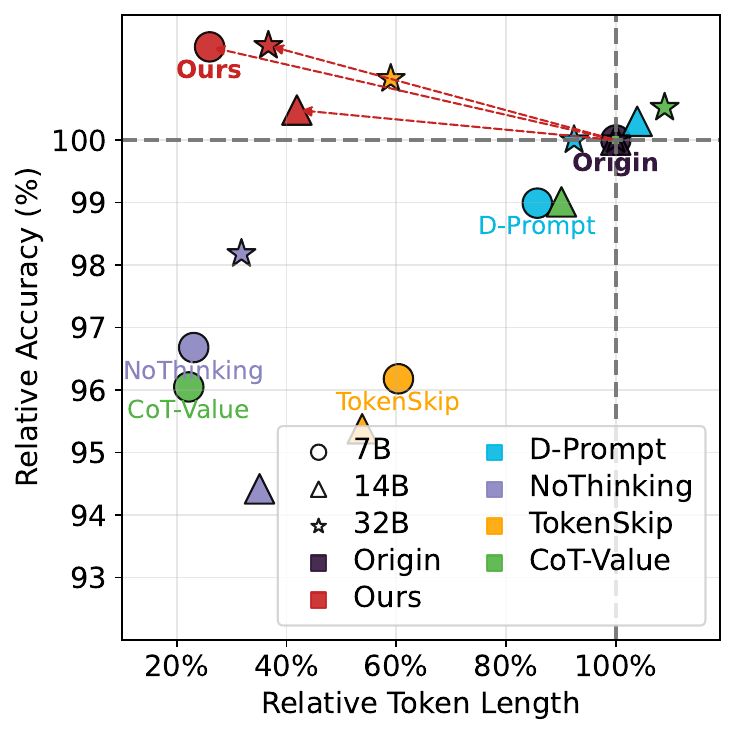}}
\subfloat[MATH]{\includegraphics[width = 0.235\textwidth]{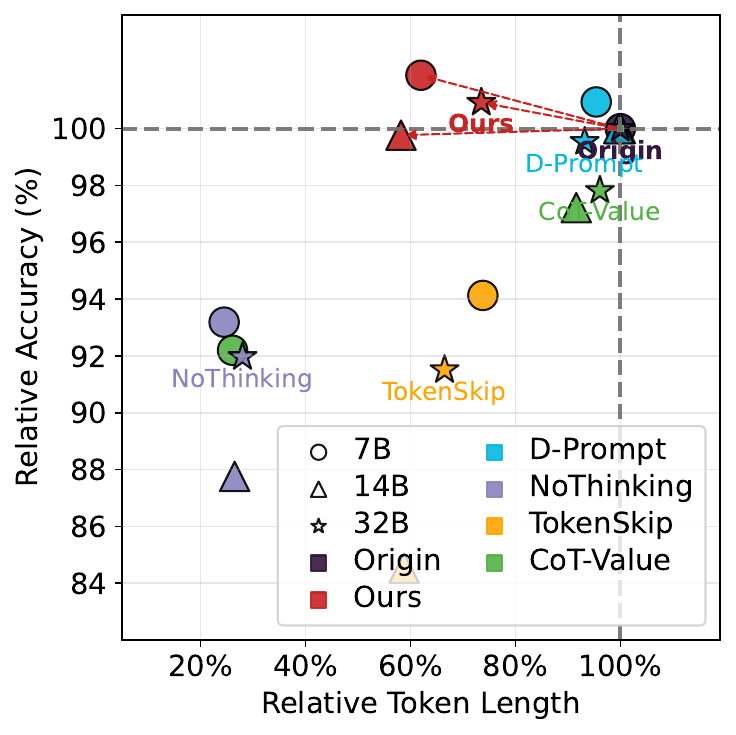}}
\caption{Comparison with baselines on R1-Distill-Qwen series models. Relative accuracy and token length of original LRM are set as 100\%.}
\label{fig:baselien_comparison}
\end{figure}

\begin{figure}[t]
\centering
\includegraphics[width = 0.475\textwidth]{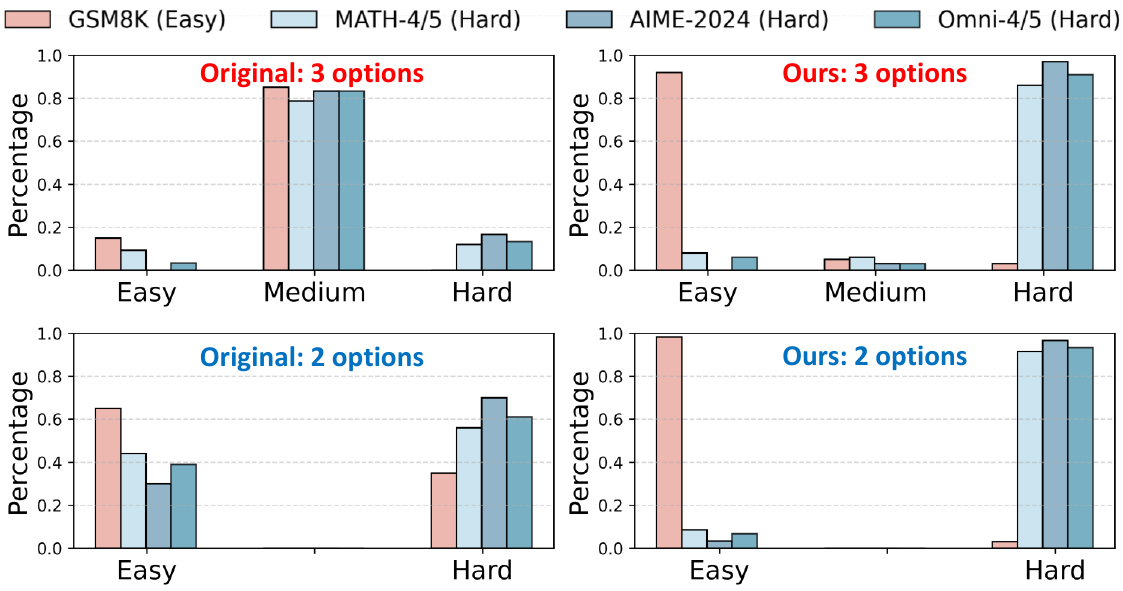}
\caption{LRM's difficulty cognition. LRM is prompted to assess the difficulty level of questions from the relatively straightforward GSM8K~\cite{cobbe2021trainingverifierssolvemath} and highly complex tasks (MATH~\cite{hendrycksmath2021}, AIME2024 and OmniMath~\cite{gao2024omnimathuniversalolympiadlevel}). Our method substantially reduces difficulty cognition conflation, consistently across both 3-option (e.g., Easy, Medium, Hard) and 2-option (e.g., Easy, Hard) setups.}
\label{fig:difficulty-cognition}
\end{figure}

Inspiration for a solution lies in the human cognitive science: the dual-process theory of human reasoning. This theory posits a cognitive architecture composed of two distinct but complementary systems: a fast, concise, and intuitive \textit{System 1}, and a slow, deliberate, and analytical \textit{System 2}~\cite{kahneman2011thinking, evans2013dual}.
Before engaging in problem-solving, humans often perform an initial assessment of a task's difficulty, allowing them to allocate cognitive resources appropriately between these two systems~\cite{efklides2011interactions, flavell1979metacognition}. For straightforward tasks, reliance on \textit{System 1} suffices, whereas complex problems necessitate the engagement of \textit{System 2}. 
Our empirical analysis (Fig.~\ref{fig:difficulty-cognition}) reveals that they approach disparate tasks—from the straightforward (e.g., GSM8K) to the highly complex (e.g., AIME2024)—with a monolithic and inefficient reasoning strategy. These limitations highlight the need for LRMs to develop autonomous difficulty cognition—to ``\textit{think how to think}''.

In this paper, we propose \textbf{T}hink-\textbf{H}ow-to-\textbf{T}hink (\textbf{TH2T}), a novel two-stage fine-tuning strategy that progressively inspires LRMs difficulty cognition and redundancy cognition. In contrast to prompt-based input guidance for instruction adherence, our approach emphasizes output-level interventions, directly intervening inside the reasoning trajectory to establish a self-hypnosis mechanism. 
In the first stage, we design \textbf{Difficulty Hypnosis (DH)} to stimulate the model’s awareness of task difficulty. DH introduces difficulty cues into output prefixes, which act as guiding signals for global, prospective strategy selection. With DH supervision from a hybrid dataset featuring both concise and elaborate reasoning trajectories, the model calibrates its sensitivity to task difficulty and adaptive control of reasoning depth, ultimately curtailing the impulse to overthink on straightforward problems through structured cognitive self-guidance.
In the second stage, we design \textbf{Redundancy Hypnosis (RH)} to cultivate the model’s awareness of redundant reasoning. RH specifically targets fine-grained inference patterns such as redundant yet logically complete structures (e.g., reflections or double-checks) and terminal reasoning loops frequently observed in complex task generation. By embedding redundancy cues into these in-progress structures, RH provides local, retrospective signals for behavior correction, enabling the model to skip or truncate superfluous steps. In this way, the model strengthens its redundancy cognition and reduces superfluous reasoning detours. Overall, our contributions can be outlined as follows:
\begin{itemize}
\item We reveal that current LRMs fall short of perceiving task complexity. To mitigate overthinking, we propose enhancing LRMs with the ability to cognize and adapt to varying levels of problem difficulty autonomously. 
\item To align with human \textit{System 1} and \textit{System 2} thinking patterns, we propose a novel two-stage fine-tuning strategy with a self-hypnosis mechanism for prospective and retrospective intervention, thus progressively inspiring LRMs' difficulty and redundancy cognition.
\item Experiments demonstrate over 70\% length reduction on easy tasks and 40\% on hard tasks with maintained robust performance, as shown in Fig.\ref{fig:baselien_comparison}, exhibiting clear difficulty-aware capabilities and reduced redundancy.
\end{itemize}
\section{Literature Review}
The advent of LRM~\cite{openai2024openaio1card,guo2025deepseek,qwq32b}
enables breakthroughs in problem solving~\cite{sun2023survey, yu2024natural} with Chain-of-Thought~\cite{wei2022chain, wang2022self} and test-time scaling~\cite{snell2025scaling}. Nevertheless, this introduces the challenge of overthinking~\cite{chen2025think23overthinkingo1like,team2025kimi,sui2025stopoverthinkingsurveyefficient}.

\noindent \textbf{Prompt-based methods.}
Some works~\cite{aytes2025sketchofthoughtefficientllmreasoning, ding2024breakchainlargelanguage, ning2024skeletonofthought,han-etal-2025-token,xu2025chain,lee2025llmscompresschainofthoughttoken} instruct LRMs by various prompts to respond concisely. Specifically, \citet{renze2024benefits} explicitly prompts LLMs to perform concise step-by-step reasoning. SoT~\cite{ning2024skeletonofthought} combines reasoning paradigms with linguistic constraints to minimize token usage. Token-Budget~\cite{han-etal-2025-token} establishes a self-estimated token budget as a constraint. \citet{ge2025innatereasoningenoughincontext} regulates thinking token distribution by few-shot in-context learning.

\noindent \textbf{Output-based methods.}
Implicit-based methods~\cite{deng2024explicitcotimplicitcot, hao2024traininglargelanguagemodels, saunshi2025reasoning, cheng2024compressedchainthoughtefficient, shen2025efficientreasoninghiddenthinking} propose to internalize reasoning steps within the latent representation, thereby shifting the cognitive process from the explicit language space to the implicit hidden space. ~\citet{saunshi2025reasoning} simulates multiple steps of CoT via a looped transformer layer to deepen the reasoning process internally. Another line of work focuses on dynamic execution. ~\citet{sun2024fast, xie2023self, liao2025rewardguided} optimize Best-of-N (BoN) decoding with speculation and rejection. ~\citet{li2024escape, yang2025dynamicearlyexitreasoning,wang2025sampling} explores potential early exit point. NoThinking~\cite{ma2025reasoningmodelseffectivethinking} prompts reasoning models to directly output final answers without thinking.

\noindent \textbf{Model-based methods.}
 Fine-tuning offers a direct post-training strategy~\cite{yang2502towards}. CoT-Value~\cite{ma-etal-2025-cot} learns a controllable direction in the parameter space to steer generation length. C3oT~\cite{kang2025c3ot} employs GPT-4 as a compressor while retaining key information. TokenSkip~\cite{xia2025tokenskip} fine-tunes the model to prune non-essential tokens during inference. Alternatively, reinforcement learning provides another vulnerable and expensive paradigm, often by incorporating length penalties to balance performance and efficiency~\cite{arora2025traininglanguagemodelsreason, team2025kimi, aggarwal2025l1controllinglongreasoning}. O1-pruner~\cite{luo2025o1prunerlengthharmonizingfinetuningo1like} integrates a Length-Harmonizing Reward into a PPO-style loss. DAST~\cite{shen2025dastdifficultyadaptiveslowthinkinglarge} employs SimPO~\cite{meng2024simpo} on a constructed length-preference dataset, while AdaR1~\cite{luo2025adar1hybridcotbileveladaptive} optimizes a merged hybrid long/short CoT model using DPO~\cite{rafailov2023direct}. 

\noindent \textbf{Limitations and our approach.}
Overall, existing approaches to efficient reasoning largely operate under a monolithic paradigm of length compression, which introduces critical limitations: 1) Prompt-based methods exhibit unreliability of instruction following and task cognition; 2) Implicit output-based methods suffer from performance degradation and compromised interpretability; 3) Model-based approaches are constrained by a behavioral rigidity that hinders adaptability.
In contrast, we reframe overthinking not just as a behavioral flaw, but as a fundamental cognitive deficit, since length is not an accurate proxy for difficulty. Our approach instills a dual-level awareness: a global, prospective cognition of task difficulty to guide high-level strategy, and a local, retrospective in-progress perception of redundancy for fine-grained execution. This self-aware intervention mechanism cultivates the metacognitive skill of thinking about how to think efficiently, rather than simply enforcing brevity.

\section{Preliminary and Observation}
\label{sec:Preliminary}
\subsection{Background}
\label{sec:Background}
\noindent \textbf{LLMs.}
Given a prompted question $x$, LLM generates a response $y := (y_1, \ldots, y_n)$ by iteratively predicting each response token $y_i = \text{LLM}([x, y_{<i}]) \in y$ conditioned on $x$ and generated tokens $y_{<i} := (y_1, \dots, y_{i-1})$. This is termed as System 1 thinking~\cite{li202512surveyreasoning}: 
$\text{[Prompt]} + \text{[Answer]}$, where $\text{[Prompt]}$ and $\text{[Answer]}$ are question $x$ and response $y$, which is often short and concise.

\noindent \textbf{LRMs.}
Whereas in LRMs, the generation process is split into two stages, i.e., thinking and conclusion, referred to as System 2:
$[\text{Prompt}] + \langle \text{think} \rangle [\text{Thoughts}] \langle / \text{think} \rangle + [\text{Conclusion}]$, where $\langle \text{think} \rangle$ and $\langle / \text{think} \rangle$ are two delimiters denoting the start and end of reasoning, $\text{[Thoughts]}$ and $\text{[Conclusion]}$ are detailed reasoning processes and concise answers, respectively.
Formally, $\text{[Thoughts]}$ reveals a trajectory of problem solving, known as CoT~\cite{wei2022chain, wang2022self}. At the micro level, some consecutive tokens in $\text{[Thoughts]}$ can be aggregated into short chunk $T_j$ based on semantics, logic and grammar:
\begin{equation}
\text{[Thoughts]} = T := (T_1, \ldots, T_m)
\label{eq:cot}
\end{equation}
Similarly, each reasoning chunk is autoregressively generated by conditioning on question $x$ and previous chunks: $T_i = \text{LRM}([x, T_{<i}])$. This explicit decomposition not only improves interpretability through transparent exposure of reasoning steps but also enhances the model’s capability for complex tasks. Specifically, some reflective chunks are well structured by prefix keywords such as ``\texttt{Wait}'', ``\texttt{However}'', ``\texttt{Alternatively}''~\cite{guo2025deepseek}, marking the subsequent chunks will undergo logical backtracking, self-verification, or exploration, which have been proved crucial to LRMs~\cite{li2025llmseasilylearnreason}. 
Under this setting, we define ``\textit{overthinking}'' as a phenomenon characterized by inefficiencies at both the global and local levels: globally, through an excessive number of reasoning steps ($T_i$) relative to task complexity, and locally, where individual steps ($T_i$) are elaborate yet logically superfluous, resulting in prominent overhead.
\subsection{Reasoning Process Intervention}
\label{sec:pre-experiement}
To address the overthinking issue, a straightforward approach is difficulty reminders—encouraging LRMs to self-regulate output length based on task difficulty judgment through prompts like ``\textit{Please think quickly if you find it easy}''. However, this hypothetical condition necessitates two prerequisites: accurate instruction following and robust task difficulty cognition. While prior studies~\cite{wu2025effectively, han-etal-2025-token} have highlighted the inherent challenges of the former, we have already shown that current LRMs lack cognition of problem difficulty in Fig.~\ref{fig:difficulty-cognition}. 

\begin{figure}[t]
\centering
\includegraphics[width = 0.99\linewidth]{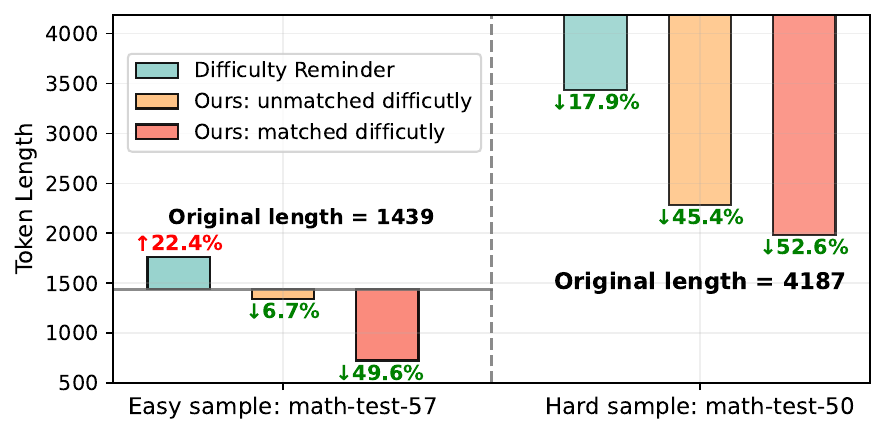}
\caption{Internal self-hypnosis vs. external difficulty reminder for response length regulation. Our internal self-hypnosis instills a robust self-regulatory mechanism, demonstrating superior effectiveness, reliability, and instruction-following capabilities compared to conventional external prompts.}
\label{fig:pre-experiemnt-effectiveness}
\end{figure}

\begin{figure*}[t!]
\centering
\includegraphics[width = 0.95\textwidth]{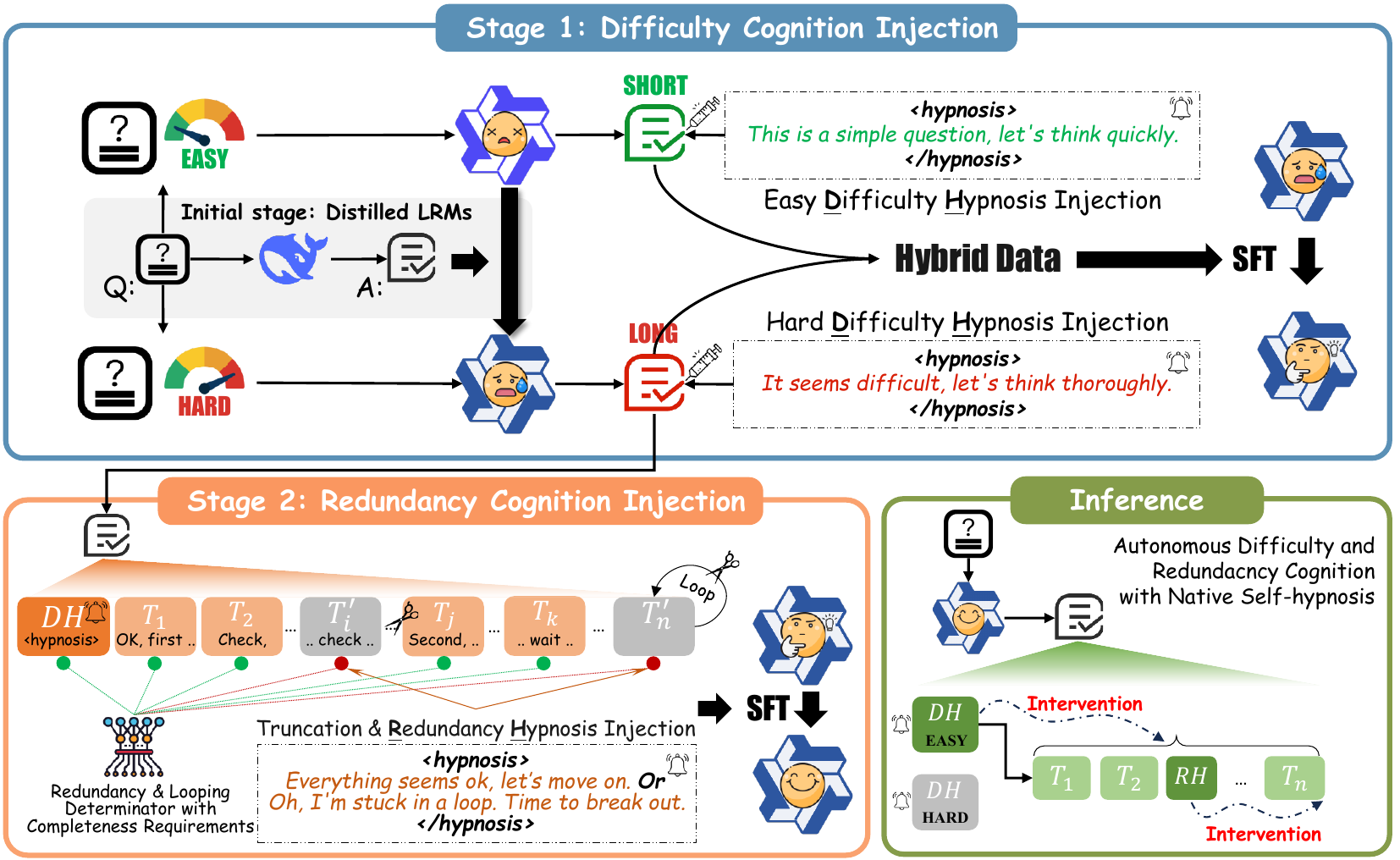}
\caption{Framework of TH2T. \textbf{Stage 1}: fine-tune with difficulty-differentiated data with injected difficulty-hypnosis, providing global, prospective signals for strategy selection. \textbf{Stage 2}: fine-tune with truncation and redundancy-hypnosis injection, providing local, retrospective signals for in-process intervention. Inference: autonomous difficulty and redundancy adaptation under the intervention of native self-hypnosis.}
\label{fig:framework}
\vspace{-2mm}
\end{figure*}

Instead, we reveal that a promising solution is transferring the intervention from the prompt to the inner reasoning process. We conduct a pre-experiment in Fig.~\ref{fig:pre-experiemnt-effectiveness}, pretending the model knows the difficulty level of samples. Given a simple problem, we change the model input from ``\textit{User: [Prompt]+[Question]. Assistant:}'' to ``\textit{User: [Prompt]+[Question]. Assistant: This is a simple question, let's think quickly.}''. While for hard problems, the predefined response prefix is ``\textit{It seems difficult, let's think thoroughly.}''. Fig.~\ref{fig:pre-experiemnt-effectiveness} shows that predefined difficulty awareness largely outperforms reminder on reducing token length, especially matched awareness (``\textit{simple, quickly}'' for easy and ``\textit{difficult, thoroughly}'' for hard is matched, the reverse is unmatched). 

We define them as self-hypnosis that acts as triggers in the output, guiding the subsequent inference trajectory. Compared to explicit prompts, these built-in triggers offer three key advantages: 1) native flexibility; 2) robustness; 3) superior instruction-following capabilities. In Sec.\ref{sec:method}, we utilize the self-hypnosis mechanism to guide LRMs to generate native interventional triggers of difficulty and redundancy cognition, autonomously.

\section{Methodology}
\label{sec:method}
As discussed in Sec.\ref{sec:Background}, the challenge of overthinking stems from failures in both high-level strategy selection and fine-grained step execution. We ground our method in the dual-process theory of cognition, i.e., intuitive System 1 and analytical System 2. 
The strategy calls for a global, prospective intervention applied pre-generation, while the execution demands a local, retrospective mechanism for in-progress correction and refinement.
Since they are complementary and distinct systems, we propose a two-stage fine-tuning framework \textbf{T}hink-\textbf{H}ow-to-\textbf{T}hink (\textbf{TH2T}) for mitigating overthinking, as shown in Fig.~\ref{fig:framework}.

\subsection{Stage 1: Difficulty Cognition Injection}
\label{sec:stage_1}
The foundational principle of our approach is to cultivate the model's capacity for task appraisal, which is a metacognitive prerequisite for adaptive reasoning. We define two datasets: $\smash{\mathcal{Q}_0 = \{ q_i^0 \mid i = 0, 1, 2, \dots, N \}}$ for easy problems, and $\smash{\mathcal{Q}_1 = \{ q_j^1 \mid j = 0, 1, 2, \dots, M \}}$ for difficult ones. Two models from the same series were selected to ensure consistency in response: $\theta_S$ for basic LLM with prompted Short-CoT capability and $\theta_L$ for advanced LRM. For each problem in $\mathcal{Q}_0$, the responses are recorded as:
\begin{equation}
\mathcal{P}_{0}^{S} = \{ q_i^0 \in \mathcal{Q}_0 \mid \mathbbold{1}[\theta_S(q_i^0)]\}
\label{eq:P0S}
\end{equation}
\begin{equation}
\mathcal{P}_{0}^{L} = \{ q_i^0 \in \mathcal{Q}_0 \mid \mathbbold{1}[\theta_L(q_i^0)]\}
\label{eq:P0L}
\end{equation}
where $\mathbbold{1}[\theta_S(q_i^0)]$ means the response of $\theta_S$ on $q_i^0$ is correct. $\theta_S$ can replicate the majority correct responses of $\theta_L$ on $\mathcal{Q}_0$, which means: ${\frac{|\mathcal{P}_{0}^{S} \cap \mathcal{P}_{0}^{L}|}{|\mathcal{P}_{0}^{L}|} \approx 1}$, wherein significant additional overhead yields limited performance gain. This motivates us to preserve $\theta_S$'s potential for efficiency on $\mathcal{Q}_0$ with satisfactory performance. The target reasoning pattern of $\theta_L$ entails concise responses on $\mathcal{Q}_0$ and detailed responses on $\mathcal{Q}_1$. To align with this objective, we construct a draft dataset $\mathcal{D}=\mathcal{D}_{0} \cup \mathcal{D}_{1}$ grounded in difficulty differentiation:
\begin{equation}
\mathcal{D}_{0} = \{ (q_i^0, \theta_S(q_i^0)) \mid q_i^0 \in \mathcal{P}_{0}^{S} \}
\label{eq:D0}
\end{equation}
\begin{equation}
\mathcal{D}_{1} = \{ (q_j^1, \theta_L(q_j^1)) \mid q_j^1 \in \mathcal{P}_{1}^{L} \}
\label{eq:D1}
\end{equation}
where $\mathcal{P}_{1}^{L} = \{ q_i^1 \in \mathcal{Q}_1 \mid \mathbbold{1}[\theta_L(q_j^1)]\}$. $\mathcal{D}_{1}$ aims to maintain the performance gain of $\theta_L$ on $\mathcal{Q}_1$.

To enable $\theta_L$ to perceive problem difficulty level autonomously, we introduce the trials of Sec.~\ref{sec:pre-experiement}. We inject specific difficulty-aware cognitive Hypnosis into the prefix of answer within $\mathcal{D}$, which, as demonstrated before, act as an interventional signals for global, prospective strategy selection.
\begin{equation}
(q, \theta(q))=(q, \ \textit{H}+\theta(q)), for  \ q \in \mathcal{D}
\end{equation}
If $q \in \mathcal{D}_{0}$, the \textbf{Difficulty Hypnosis} $\textit{H}$ is defined as follows: ``\textit{$\langle$hypnosis$\rangle$ This is a simple question, let's think quickly.$\langle$/hypnosis$\rangle$}''. If $q \in \mathcal{D}_{1}$, then ``\textit{$\langle$hypnosis$\rangle$It seems difficult, let's think thoroughly.$\langle$/hypnosis$\rangle$}''. The markers ``\textit{$\langle$hypnosis$\rangle$}'' and ``\textit{$\langle$/hypnosis$\rangle$}'' are indicators for inserted difficulty hypnosis. The injected hypnosis is recognized as an embedded difficulty label to train $\theta_L$ to generate native difficulty cognition, stimulating a concise thinking pattern on easy tasks. The execution of stage 1 is outlined in the upper section of Fig.~\ref{fig:framework}.

\subsection{Stage 2: Redundancy Cognition Injection}
We then extend difficulty-hypnosis to redundancy-hypnosis, guiding $\theta_L$ to identify redundant structures within the reasoning process on hard tasks. 

We observe severe redundant structures in $\theta_L$'s response on hard tasks, including superfluous reflection and check, along with a remarkably high proportion of reasoning loops in incorrect answers. According to Eq.~(\ref{eq:cot}), the response of  $q_j^1 \in \mathcal{D}_{1}$ can be aggregated into chunks based on logical semantic or grammatical rules like ``\texttt{\textbackslash n\textbackslash n}'' delimiter. The redundancy issue can be formalized as $\theta_L(q_j^1)$ :
\begin{equation}
(H, T_1, ...,T_{i}^{'}, ...,T_k, ...,T_{n}^{'},T_{n}^{'},T_{n}^{'}, ... )
\label{eq:redundancy}
\end{equation}
where $H$ is the difficulty-hypnosis introduced in Stage 1, $T_1,T_K$ are valid reasoning chunks to be kept. $\smash{T_{i}^{'}}$ is the superfluous reflection structure inside the reasoning process without effective contribution to the reasoning process. $\smash{T_{n}^{'}}$ is a similar reflection structure; nonetheless, it primarily manifests in the final phase of the reasoning, and the model enters an unproductive loop (trap), relentlessly iterating until $\theta_L$ reaches the maximum output length. This practice severely degrades inference efficiency, as latency often scales super-linearly with sequence length, and the vast majority of problems do not require such exhaustive thought.

We define a \textbf{Determinator} to judge whether some $T_{m} \in \theta_L(q_j^1)$ is logically equivalent to $\smash{T_{i}^{'}}$ or $\smash{T_{n}^{'}}$ (i.e., redundant or looping). If yes, we inject redundancy hypnosis and truncate subsequent content. In case of $\smash{\textit{Determinator}(T_{m}) \equiv T_{i}^{'}}$, original $\smash{\theta_{L}(q_j^1)}$ is refined as:
\begin{equation}
\smash{\theta_{L}^{'}(q_j^1)=(H, T_1, T_2, ...,T_{m-1},\bar{H}_1)}
\label{eq:redundancy_hypnosis}
\end{equation}
\noindent wherein the \textbf{Redundancy Hypnosis} $\smash{\bar{H}_1}$ is defined as follows: ``\textit{$\langle$hypnosis$\rangle$ Everything seems ok, let's move on.$\langle$/hypnosis$\rangle$}''.
We use GPT-4~\cite{openai2024gpt4technicalreport} to implement the determinator. 
In case of $\smash{\textit{Determinator}(T_{m}) \equiv T_{n}^{'}}$, the training set lacks a sufficient number of loop structures; meanwhile, excessively deferred looping hinders the generation of shorter-length samples. Thus, we select intermediate chunk with words like ``\texttt{Wait}'', ``\texttt{However}'', ``\texttt{Alternatively}'' and iterate it as a simulation of looping, followed by hypnosis injection and truncation:
\begin{equation}
\smash{\theta_{L}^{'}(q_j^1)=(H, T_1, T_2, ..., T_{m}, T_{m}, T_{m},\bar{H}_2)}
\label{eq:looping_hypnosis}
\end{equation}
\noindent wherein the \textbf{looping hypnosis} $\smash{\bar{H}_2}$ is defined as follows: ``\textit{$\langle$hypnosis$\rangle$ Oh, I'm stuck in a loop. Time to break out. $\langle$/hypnosis$\rangle$}''. 
To determine the optimal timing for truncation, we leverage DeepSeek-R1 to justify the earliest point—a condition we term the completeness requirement—where the preceding content becomes sufficient to derive the correct answer. 
The choice of iterate number $m$ can be random, and we also try a probe-like approach~\cite{zhang2025reasoning}. The last-layer hidden states of the selected chunk tail are extracted to train a simple two-layer MLP as Determinator. Experiments indicate it yields similar results to the random approach, so we prefer the latter. 
Notably, we adopt truncation instead of mere substitution, as truncation allows for the construction of shorter samples. Meanwhile, generation continuity is ensured by remaining untruncated samples in the dataset, since we only apply Stage 2 to part of the samples in $\mathcal{D}_{1}$.
\section{Experiments}
\label{sec:experiments}
\subsection{Experiment Setup}
\label{sec:setup}
\noindent \textbf{Models.}
We do experiments on Long-CoT models, including DeepSeek-R1-Distilled-Qwen-2.5-7B/14B/32B and -Llama-8B~\cite{guo2025deepseek}. During data construction, the adopted Short-CoT model are Qwen2.5-7B/14B/32B~\cite{qwen2025qwen25technicalreport} and Llama-3.1-8B-Instruct~\shortcite{grattafiori2024llama3herdmodels}. They are $\theta_L$ and $\theta_S$, respectively.

\noindent \textbf{Datasets.}
For dataset construction, we mainly adopt mathematical reasoning tasks, which are friendly for difficulty quantization. In stage 1, the GSM8K and MATH datasets are combined after filtering out errors and injecting difficulty hypnosis. In stage 2, redundancy and looping hypnosis are injected into difficult samples. The composition ratio of this heterogeneous dataset is 2:1:1, yielding $8k$ samples. We evaluate on 4 mathematical tasks (GSM8K, MATH, AIME2024 and Omni-MATH) and 1 QA benchmark GPQA~\cite{rein2024gpqa}. Metrics include accuracy, average number of output tokens, and inference latency per sample. We also report the success rate of the difficulty awareness, redundancy, and looping optimization.

\noindent \textbf{Baselines.}
We compare \textbf{TH2T} with the following methods: 
(1) \textbf{D-Prompt} (prompt-based) directly regulates response length based on self-judgment of task difficulty via the Difficulty Reminder in Prompt (in Sec.\ref{sec:pre-experiement}). 
(2) \textbf{NoThinking}~\cite{ma2025reasoningmodelseffectivethinking} (output-based) bypasses the thinking process via prompting for all difficulty-level questions. 
(3) \textbf{Tokenskip}~\cite{xia2025tokenskip} (model-based) fine-tunes the model to prune non-essential tokens during inference, regardless of task difficulty. 
(4) \textbf{CoT-Valve}~\cite{ma-etal-2025-cot} (model-based) learns a controllable direction in the parameter space to steer generation length based on task difficulty. 
(5) \textbf{AdaCtrl}\footnote{We re-implemented all baselines except AdaCtrl (citing its original 7B results due to reproduction complexity).} ~\cite{huang2025adactrl} (model-based) uses reinforcement learning and allocates reasoning budgets based on self-assessed problem difficulty, which is more of a label-driven distillation.

\begin{table*}[t]   \scriptsize
\centering
\setlength{\tabcolsep}{1.8mm}
\begin{tabular}{@{}l crr crr crr crr crr@{}}
\toprule
\multirow{2}{*}{\textbf{Methods}} & \multicolumn{3}{c}{\textbf{GSM8K}}  & \multicolumn{3}{c}{\textbf{MATH-500}} & \multicolumn{3}{c}{\textbf{AIME2024}} & \multicolumn{3}{c}{\textbf{OmniMath}} & \multicolumn{3}{c}{\textbf{GPQA}}\\ 
\cmidrule(lr){2-4} \cmidrule(lr){5-7} \cmidrule(lr){8-10} \cmidrule(lr){11-13} \cmidrule(lr){14-16}
  &Acc. &Len. &Reduc. &Acc. &Len. &Reduc. &Acc. &Len. &Reduc. &Acc. &Len. &Reduc. &Acc. &Len. &Reduc.  \\ 
\midrule
\multicolumn{16}{c}{\textit{DeepSeek-R1-Distill-Qwen-7B}} \\ \cmidrule{1-16}
{Original}  &91.51 &1057 &  &85.2 &2857 & &\textbf{50.0} &10570 & &45.0 &5736 & &\underline{32.8} &5349 & \\ 
\hdashline[1pt/2pt]
{D-Prompt} & 90.59 &906 &$-$14.3\% &\underline{86.0} &2726 &$-$\ \ 4.6\% &46.7 &10209 &$-$\ \ 3.4\% &45.0 &5398 &$-$\ \ 5.9\% &30.3 &5117 &$-$\ \ 4.3\%\\
{NoThinking} & 88.47 &243 &$-$76.9\% &79.4 &699 &$-$75.5\%  &40.0 &4134 &$-$60.9\% &40.0 &2083 &$-$63.7\% &29.3 &2084 &$-$61.0\%\\  
{TokenSkip} &88.01 &638 &$-$39.6\% &80.2 &2109 &$-$26.2\%  &40.0 &5559 &$-$47.4\% &38.3 &3461 &$-$39.7\% &28.8 &4001 &$-$25.2\%\\
{CoT-Valve} &87.90 &235 &$-$77.8\% &78.6 &747 &$-$73.9\%  &43.3 &5871 &$-$44.5\% &41.7 &3311 &$-$42.3\% &28.3 &3226 &$-$39.7\%\\

{AdaCtrl} &90.98 &349 &\multicolumn{1}{c}{-} &74.0 &3196 &\multicolumn{1}{c}{-} &21.3 &16889 &\multicolumn{1}{c}{-} &\multicolumn{1}{c}{-} &\multicolumn{1}{c}{-} &\multicolumn{1}{c}{-} &\multicolumn{1}{c}{-} &\multicolumn{1}{c}{-} &\multicolumn{1}{c}{-} \\

\hdashline[1pt/2pt]
\rowcolor{mylightgray} (\textbf{w/o D.H.})  & 90.83 &503 &$-$47.4\% &85.2 &1906 &$-$33.2\%  &\textbf{50.0} &8213 &$-$22.3\% &\underline{46.7} &3571 &$-$37.7\% &\textbf{33.3} &4372 &$-$18.3\% \\

\rowcolor{mylightgray} (\textbf{w/o R.H.})  & \underline{91.96} &397 &$-$62.4\% &\underline{86.0} &2265 &$-$20.7\%  &\underline{46.7} &8684 &$-$17.8\% &\textbf{48.3} &4247 &$-$26.0\% &\underline{32.8} &4481 &$-$16.2\%\\

\rowcolor{mygray} \textbf{TH2T}  &\textbf{92.87} &275 &$-$74.0\% &\textbf{86.8} &1772 &$-$38.0\%  &\textbf{50.0} &7490 &$-$29.1\% &\textbf{48.3} &3377 &$-$41.1\% &\underline{32.8} &4166 &$-$22.1\%\\ 
\toprule
\multicolumn{16}{c}{\textit{DeepSeek-R1-Distill-Qwen-14B}} \\ \cmidrule{1-16}
{Original} &93.63 &584 & &86.6 &2306 & &\textbf{50.0} &9900 & &46.7 &5373 & &50.0 &3847 & \\ 
\hdashline[1pt/2pt]
{D-Prompt} &93.91 &606 &$+$\ \ 3.9\% &86.6 &2297 &$-$\ \ 0.3\% &50.0 &10408 &$+$\ \ 5.1\% &41.7 &4994 &$-$\ \ 7.1\% &49.0 &3946 &$+$\ \ 2.6\% \\
{NoThinking} &88.41 &206 &$-$64.9\% &76.0 &612 &$-$73.5\% &43.3 &3735 &$-$62.3\% &43.3 &1944 &$-$63.8\% &45.5 &1454 &$-$62.2\% \\ 
{TokenSkip} &89.30 &314 &$-$46.2\% &73.2 &1356 &$-$41.2\% &\underline{46.7} &6772 &$-$31.6\% &40.0 &3158 &$-$41.2\% &46.9 &2465 &$-$35.9\% \\
{CoT-Valve} &92.70 &526 &$-$\ \ 9.9\%  &84.2 &2111 &$-$\ \ 8.4\% &40.0 &7027 &$-$29.0\% &41.7 &2751 &$-$48.8\% &46.5 &2872 &$-$25.3\%  \\
\hdashline[1pt/2pt]
\rowcolor{mylightgray} (\textbf{w/o D.H.})  &93.69 &338 &$-$42.2\% &86.2 &1549 &$-$32.8\% &\underline{46.7} &7807 &$-$21.1\% &\underline{46.7} &3745 &$-$30.3\% &\underline{50.5} &2208 &$-$42.6\% \\

\rowcolor{mylightgray} (\textbf{w/o R.H.})  &93.47 &299 &$-$51.2\% &85.8 &1706 &$-$26.0\% &\textbf{50.0} &8156 &$-$17.6\% &\textbf{48.3} &3858 &$-$28.2\% &\underline{50.5} &2463 &$-$36.0\% \\

\rowcolor{mygray} \textbf{TH2T}  &\textbf{94.08} &246 &$-$58.1\% &\textbf{86.4} &1343 &$-$41.8\% &\textbf{50.0} &7140 &$-$27.9\% &\textbf{48.3} &3486 &$-$35.1\% &\textbf{51.0} &1901 &$-$50.6\%   \\ 
\toprule
\multicolumn{16}{c}{\textit{DeepSeek-R1-Distill-Qwen-32B}} \\ \cmidrule{1-16}
{Original}  &95.60 &717 & &87.2 &2357 & &\textbf{60.0} &9605 & &\textbf{50.0} &5540 & &\textbf{52.0} &4384 &  \\ 
\hdashline[1pt/2pt]
{D-Prompt} &95.60 &663 &$-$\ \ 7.6\% &86.8 &2198 &$-$\ \ 6.8\% &56.7 &9445 &$-$\ \ 1.7\% &48.3 &5135 &$-$\ \ 7.3\% &49.5 &3877 &$-$11.6\%  \\
{NoThinking} &93.86 &228 &$-$68.2\% &80.2 &659 &$-$72.0\% &53.3 &3208 &$-$66.6\% &45.0 &2146 &$-$61.3\% &50.5 &1627 &$-$62.9\%   \\ 
{TokenSkip} &96.54 &423 &$-$41.0\% &79.8 &1567 &$-$33.5\% &50.0 &5856 &$-$39.0\% &43.3 &3730 &$-$32.7\% &47.4 &2874 &$-$34.4\% \\
{CoT-Valve} &96.10 &781 &$+$\ \ 8.9\% &85.3 &2263 &$-$\ \ 3.9\% &53.3 &7997 &$-$16.7\% &43.3 &4663 &$-$15.8\% &49.0 &3550 &$-$19.0\% \\
\hdashline[1pt/2pt]
\rowcolor{mylightgray} (\textbf{w/o D.H.})  &96.08 &361 &$-$49.7\% &87.4 &1863 &$-$20.9\% &\underline{56.7} &7737 &$-$19.4\% &\underline{48.3} &3432 &$-$38.1\% &\underline{51.5} &2513 &$-$42.7\%  \\

\rowcolor{mylightgray} (\textbf{w/o R.H.})  &95.75 &323 &$-$55.0\% &86.8 &2114 &$-$10.3\% &\textbf{60.0} &8155 &$-$15.1\% &45.0 &3644 &$-$34.2\% &51.0 &2809 &$-$35.9\%  \\

\rowcolor{mygray} \textbf{TH2T}  & \textbf{97.04} &263 &$-$63.3\% &\textbf{88.0} &1733 &$-$26.5\% &\underline{56.7} &7004 &$-$27.1\% &\underline{48.3} &2998 &$-$45.9\% &\textbf{52.0} &2367 &$-$46.0\%  \\

\bottomrule
\end{tabular}
\caption{Main results of R1-Distilled 7/14/32B models on 5 benchmarks. We report accuracy, average token length and length reduction ratio. \textbf{D.H.} denotes Difficulty Hypnosis $H$, \textbf{R.H.} includes Redundancy Hypnosis $\bar{H}_1$ and Looping Hypnosis $\bar{H}_2$. (Reduction abbreviated as ``Reduc'', \textbf{Bold}: best result, \underline{underline}: second-best result)}
\label{tab:main}
\vspace{-1mm}
\end{table*}
\begin{table}[t!] \scriptsize
\centering
\setlength{\tabcolsep}{2.0mm}
\begin{tabular}{lcrrrr}
\toprule
\multirow{2}{*}{\textbf{Bench}} & \multirow{2}{*}{\textbf{Difficulty lv.}} & \multicolumn{2}{c}{\textbf{Cognition Acc.}} & \multicolumn{2}{c}{\textbf{Confidence}}\\ 
\cmidrule(lr){3-4} \cmidrule(lr){5-6}
                           &               & Original & TH2T & Original & TH2T\\ 
\midrule
\multirow{1}{*}{GSM8K}     & all easy      & 14.9\%  & \textbf{98.3\%} &15.46 &\textbf{17.19}  \\
\midrule
\multirow{2}{*}{MATH}      & lv.1-2 (easy) & 9.8\%   & \textbf{93.4\%} &16.09 &\textbf{16.97}  \\
                           & lv.4-5 (hard) & 12.0\%  & \textbf{91.5\%} &15.72 &\textbf{16.25}  \\
\midrule
\multirow{1}{*}{AIME2024}  & all hard      & 16.7\%  & \textbf{96.7\%} &16.64 &\textbf{16.89}  \\
\midrule
\multirow{2}{*}{OmniMath}  & lv.1-2 (easy) & 20.0\%  & \textbf{96.7\%} &16.41 &\textbf{17.05}  \\
                           & lv.4-5 (hard) & 13.3\%  & \textbf{93.3\%} &16.94 &\textbf{17.76}  \\
\bottomrule
\end{tabular}
\caption{Difficulty Cognition. We compare the correctness of difficulty assessment by R1-Distilled-7B and our TH2T-7B, further examining the first-token confidence as an internal signal for self-consistent metacognition.}
\label{tab:difficulty_awareness}
\end{table}
\subsection{Results}
\label{sec:results}
\noindent \textbf{TH2T achieves a significant reduction in computational overhead without compromising performance.}
The in-distribution performance on GSM8K and MATH is illustrated in Tab.~\ref{tab:main}. 
Key observations include: 1) TH2T exhibits minimal accuracy degradation; more precisely, most settings witness slight improvement. 
2) TH2T achieves the best reduction in reasoning length among methods with stable accuracy---up to 74\% on GSM8K and 38\% on MATH with the 7B model, translating to over 5$\times$ and 2$\times$ latency speedups, respectively.
3) Greater gains are achieved on easier tasks, validating TH2T's ability to adapt reasoning effort based on task difficulty.
4) These observations are consistent across 7B, 14B and 32B models. 
As shown in Fig.~\ref{fig:baselien_comparison}, other methods (e.g., NoThinking and CoT-Value) achieve comparable compression ratios in token length but at the cost of severe accuracy degradation (often $>$10 points), underscoring TH2T’s robustness in balanced efficiency.

\noindent \textbf{TH2T substantially strengthens the calibration of task difficulty cognition.} The entirety of GSM8K, MATH lv.1-2, and Omni-Math lv.1-2 are set as \textit{``Easy''} while the remaining are \textit{``Hard''}. 
As shown in Tab.~\ref{tab:difficulty_awareness} and Fig.~\ref{fig:difficulty-cognition}, the original LRM reveals ambiguous recognition of task difficulty (typically $<20\%$), defaulting to the Medium label in the 3-option setting and lacking discriminability in the 2-option setting. In contrast, our approach yields a clear distinction with precision above 90\%, evidencing effective calibration of task difficulty cognition enabled by the first-stage training.

\begin{figure}[t]
\centering
\subfloat[Original-7B]{\includegraphics[width = 0.235\textwidth]{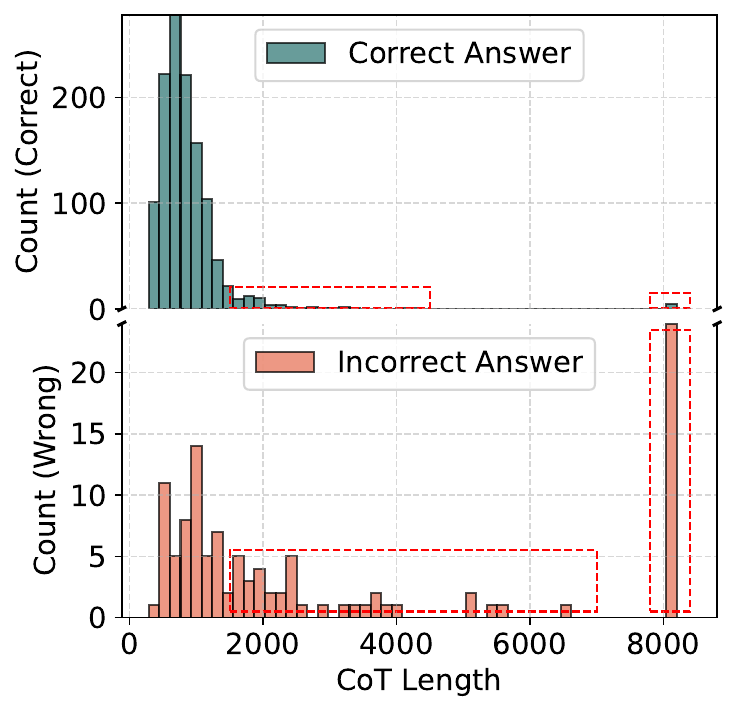}}
\subfloat[TH2T-7B]{\includegraphics[width = 0.235\textwidth]{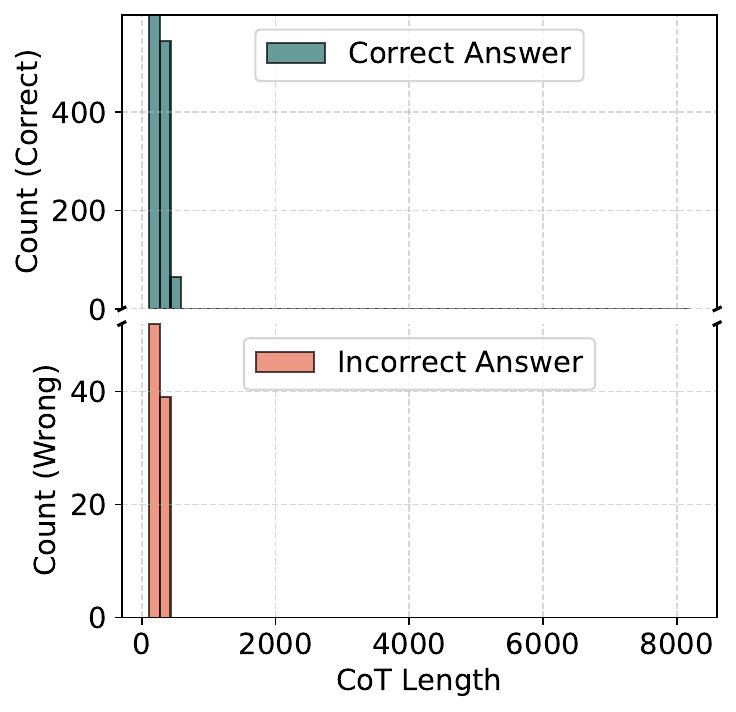}}
\caption{Statistics of length distribution on GSM8K. Our approach eliminates longer responses, especially repetitive ones that reach maximum generation length, presenting more efficiency in terms of redundancy.}
\label{fig:len_distribution_correct_incorrect}
\end{figure}

\noindent \textbf{TH2T substantially enhances cognition of redundancy.} Tab.~\ref{tab:redundancy_count} reports the average number of reflective and looping structures, comparing R1-Distilled-7B and TH2T-7B. We observe that: 1) TH2T effectively suppresses the number of reflection chunks by 10$\times$ on easy tasks and 3$\times$ on hard tasks, suggesting that extensive reflection is superfluous for strong reasoning. 2) TH2T substantially alleviates the looping problem. Notably, 59.5\% of incorrect answers by the original model on MATH contain tail recursion, which is reduced to 21.4\% under TH2T. As shown in Fig.~\ref{fig:len_distribution_correct_incorrect}(b), we observe zero loop structure in GSM8K. This distinct pattern of short-length concentration highlights the efficacy of Stage-2 of TH2T, yielding more concise and efficient reasoning structures.

\begin{table}[t]  \footnotesize
\centering
\setlength{\tabcolsep}{4pt}
\begin{tabular}{ll rr rr}
\toprule
\multirow{2}{*}{\textbf{Bench}} & \multirow{2}{*}{\textbf{Split}} & \multicolumn{2}{c}{\textbf{Reflective}} & \multicolumn{2}{c}{\textbf{Looping}} \\ 

\cmidrule(lr){3-4} \cmidrule(lr){5-6}
                       &              &Original &TH2T      &Original &TH2T \\ 
\midrule
\multirow{3}{*}{GSM8K} & Overall      &10.3  &\textbf{1.2} &2.0\% &\textbf{0.0\%}  \\
                       & --correct    &4.8   &\textbf{0.0} &0.4\% &\textbf{0.0\%}  \\
                       &--incorrect   &69.4  &\textbf{10.8} &22.3\% &\textbf{0.0\%} \\ 
\midrule
\multirow{3}{*}{MATH}  & Overall      & 27.0 &\textbf{12.3} &10.4\% &\textbf{4.0\%} \\
                       & --correct    & 14.9 &\textbf{6.8}  &1.9\%  &\textbf{1.0\%} \\
                       &--incorrect   & 96.5 &\textbf{39.1} &59.5\% &\textbf{21.4\%} \\
\bottomrule
\end{tabular}
\caption{Statistics of reflective and looping structure counts (Original R1-Distilled-Qwen-7B \textit{vs} TH2T-7B). Reflective structures include chunks containing \textit{``Wait''}, \textit{``Alternatively''}, or \textit{But''}. Looping structures are defined as extremely long answers ending with repetitive patterns up to the maximum generation length. Tasks are further divided based on correctness.}
\label{tab:redundancy_count}
\end{table}

\subsection{Analysis} 
\label{sec:analysis}
\noindent \textbf{Generalization Ability.} 
To validate the generalization on OOD tasks, we evaluate on 2 more mathematical benchmarks, i.e., AIME2024 and OmniMath, and 1 QA benchmark, i.e., GPQA. As shown in Tab.~\ref{tab:main}, consistent with the main results, our approach preserves accuracy while yielding substantial token reduction up to 20-50\%.

\noindent \textbf{Internal states validate the enhanced cognition.}
To examine whether our method instills a genuine sense of difficulty cognition, we investigate the model's internal state in Tab.\ref{tab:difficulty_awareness}. We posit that the confidence~\cite{fu2025deep} of the first token acts as an intrinsic, metacognitive signal of the model's perceived task difficulty. Our method exhibits higher confidence, aligning with a concurrent rise in difficulty assessment accuracy. This consistency serves as a clear, self-consistent internal signal, providing strong validation that the enhanced cognition is explicitly represented in internal state.

\paragraph{Ablation on two-stage design.}
We explore two more settings: 1) without Difficulty Hypnosis in Stage-1, and 2) without Redundancy Hypnosis in Stage-2, which are annotated as \textit{w/o D.H.} and \textit{w/o R.H.} in Tab.~\ref{tab:main}. We observe that D.H. introduces more token reduction on easy task GSM8K than on hard task MATH (e.g., 26.6\% \textit{vs} 4.8\%, on 7B model), while R.H. exerts a greater impact on MATH (11.6\% \textit{vs} 17.3\%). As we illustrate in Sec.\ref{sec:method}, we design Difficulty Cognition Injection Stage to build native perception on easy tasks and the Redundancy Cognition Injection Stage to mitigate non-essential structures that frequently occur in hard tasks. This phenomenon also holds for the other three datasets, demonstrating the rationality of our two-stage framework.

\begin{figure}[t]
\centering
\includegraphics[width = 0.50\textwidth]{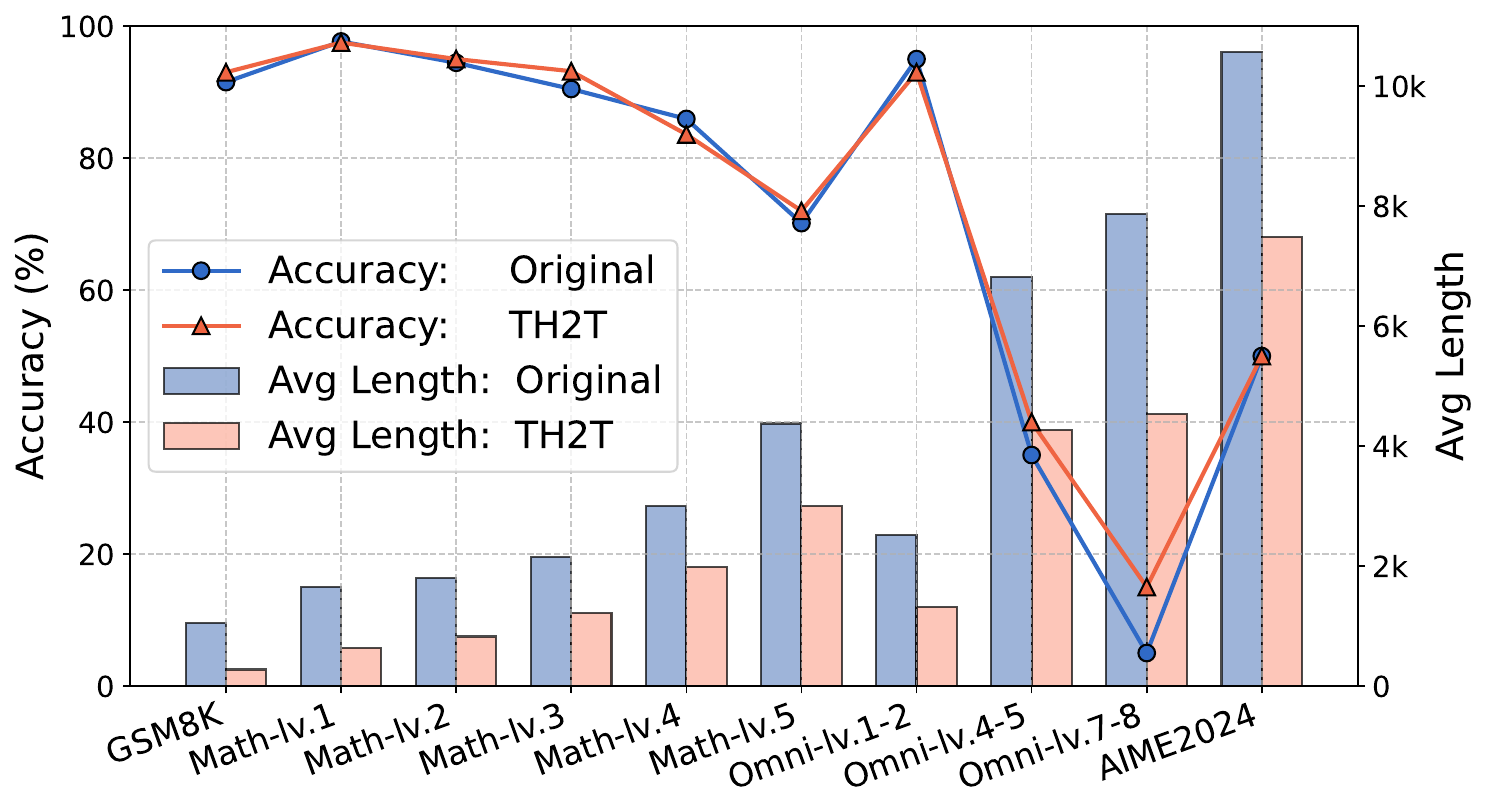}
\caption{Breakdown of the acc and average length distribution across the spectrum of difficulty levels.}
\label{fig:acc_len_on_difficulty_splits}
\vspace{-3mm}
\end{figure}

\paragraph{Acc and length across difficulty levels.}
We report a breakdown of the accuracy and average token length distribution across the complete spectrum of difficulty levels on 4 tasks in Fig.\ref{fig:acc_len_on_difficulty_splits}. 
We compare the original R1-Distilled-Qwen-7B with TH2T-7B, where higher numerical levels indicate greater question difficulty. Our observations are as follows: 1) Both models produce progressively longer and more elaborate responses with increasing difficulty. 2) Across all difficulty levels and tasks, TH2T consistently reduces response length while maintaining stable performance. 3) Length reduction is most pronounced for easier questions.

\section{Conclusion}
This paper presents TH2T, a novel two-stage fine-tuning strategy for large reasoning models (LRMs) that effectively addresses the challenge of overthinking. Our empirical analysis indicates that current LRMs are largely limited to recognizing task difficulty, exhibiting a necessity for aligning with human \textit{System 1} and \textit{System 2} thinking patterns. 
Motivated by this, TH2T progressively enables native difficulty and redundancy cognition based on a hybrid dataset and proposed self-hypnosis mechanism. To the best of our knowledge, we are the first to inject difficulty/redundancy hypnosis as built-in triggers for global, prospective, or local, retrospective interventional signals. 
Experiments demonstrate that TH2T significantly reduces inference length without compromising performance, exhibiting clear sensitivity to difficulty and redundancy.

\section*{Limitations}
In real-world cognitive settings, difficulty exists on a continuum, unlike the 2- or 3-level discretization employed in this study. However, to maintain research focus and interpretability, we adopt this simplified cognitive modeling approach.
Also, our current approach primarily focuses on standard fine-tuning using the generated hybrid dataset. While this method is effective within our current framework, it may not fully exploit the potential for optimized efficiency. Reinforcement learning presents a promising avenue for further performance gains, although it can be unstable and computationally demanding.
Finally, we acknowledge that, in theory, models of parameter scales exhibit distinct capability boundaries, which result in varying levels of task difficulty perception. At this stage, we apply a unified difficulty evaluation standard across models of different scales. Exploring the relationship between a model’s cognition and its capability boundaries is a promising research direction, potentially by dynamic correctness ratio during sampling, despite its significant computational overhead.

\bibliography{custom}

\appendix

\twocolumn[{
  \begin{center}
    \LARGE\bfseries Appendices \par
    \vspace{1em} 
  \end{center}
}]

\begin{table*}[!htbp]   \footnotesize
\centering
\begin{tabular}{@{}l cc cc cc cc cc cc@{}}
\toprule
\multirow{2}{*}{\textbf{Methods}} 
& \multicolumn{2}{c}{\textbf{GSM8K}}  
& \multicolumn{2}{c}{\textbf{MATH-500}} 
& \multicolumn{2}{c}{\textbf{AIME2024}} 
& \multicolumn{2}{c}{\textbf{OmniMath}} 
& \multicolumn{2}{c}{\textbf{GPQA}} 
& \multicolumn{2}{c}{\textbf{Average}}\\ 
\cmidrule(lr){2-3} \cmidrule(lr){4-5} \cmidrule(lr){6-7} \cmidrule(lr){8-9} \cmidrule(lr){10-11} \cmidrule(lr){12-13}
  & Acc & Len & Acc & Len & Acc & Len & Acc & Len & Acc & Len & Acc & Len \\ 
\midrule
\multicolumn{13}{c}{\textit{DeepSeek-R1-Distill-Llama-8B}} \\ 
\cmidrule{1-13}
{Original}      &75.36 &260 &61.8 &701 &3.3  &1378 &26.67 &844 &34.34 &704  &40.29 &777 \\ 
{D-Prompt}      &75.14 &261 &\textbf{62.2} &596 &6.6  &1100 &28.33 &808 &34.34 &669  &41.32 &687 \\
{NoThinking}    &75.24 &257 &\textbf{62.2} &580 &6.6  &1019 &28.33 &714 &32.32 &619  &40.94 &638 \\
{\textbf{TH2T}} &\textbf{75.58} &\textbf{245} &61.6 &\textbf{498} &\textbf{13.3} &\textbf{861}  &\textbf{40.00} &\textbf{561} &\textbf{40.91} &\textbf{526}  &\textbf{46.28} &\textbf{538} \\
\bottomrule
\end{tabular}
\caption{Results of R1-Distilled-Llama-8B model on 4 mathematical benchmarks (i.e., GSM8K, MATH, AIME2024, and OmniMath) and 1 QA benchmark (i.e., GPQA). We report accuracy, average token length for comparison.}
\label{tab:main-llama}
\end{table*}

\begin{figure*}[!htbp]
\centering
\subfloat[Acc on GSM8K]{\includegraphics[width = 0.35\textwidth]{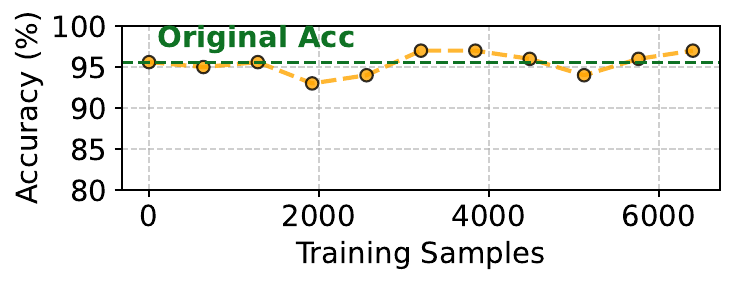}}
\subfloat[Acc on MATH]{\includegraphics[width = 0.35\textwidth]{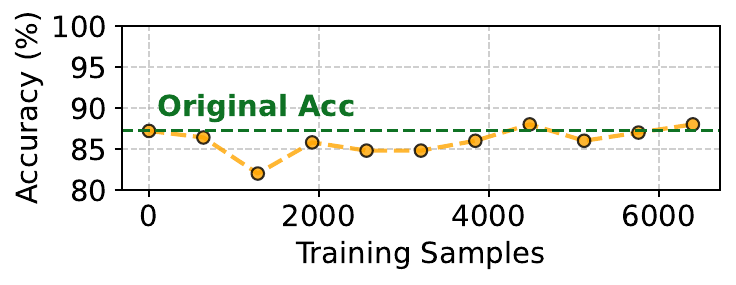}}
\hfill
\subfloat[Token length on GSM8K]{\includegraphics[width = 0.35\textwidth]{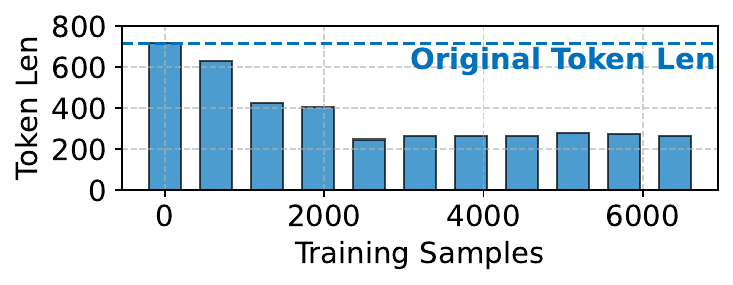}}
\subfloat[Token length on MATH]{\includegraphics[width = 0.35\textwidth]{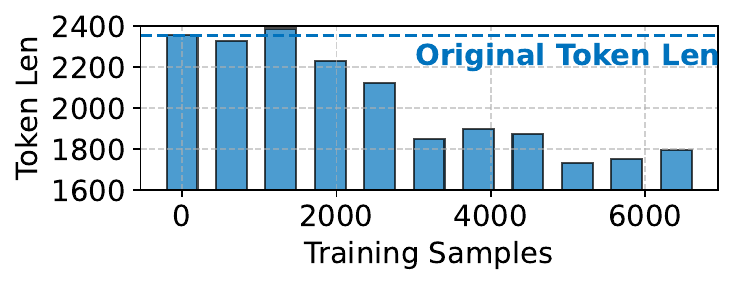}}
\caption{Dynamics of accuracy and average token length during the training process (R1-Distill-Qwen-32B). We find: 1) As training steps increase, accuracy initially declines, then returns to the original level, with occasional positive gains during the process. 2) For hard tasks like MATH, the response length exhibits a reverse increase after exceeding a certain number of steps.}
\label{fig:train_process}
\end{figure*}

\begin{figure*}[!htbp]
\centering
\subfloat[on GSM8K benchmark]{\includegraphics[width = 0.35\textwidth]{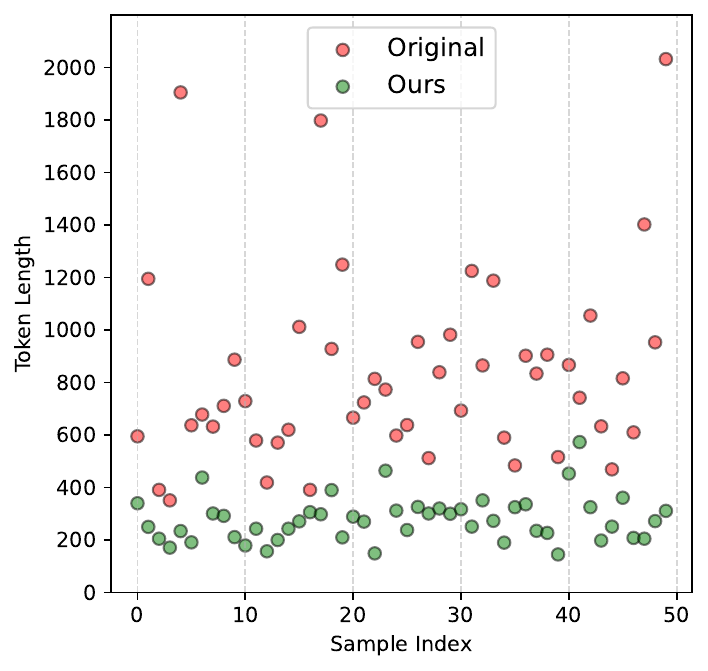}}
\subfloat[on MATH benchmark]{\includegraphics[width = 0.35\textwidth]{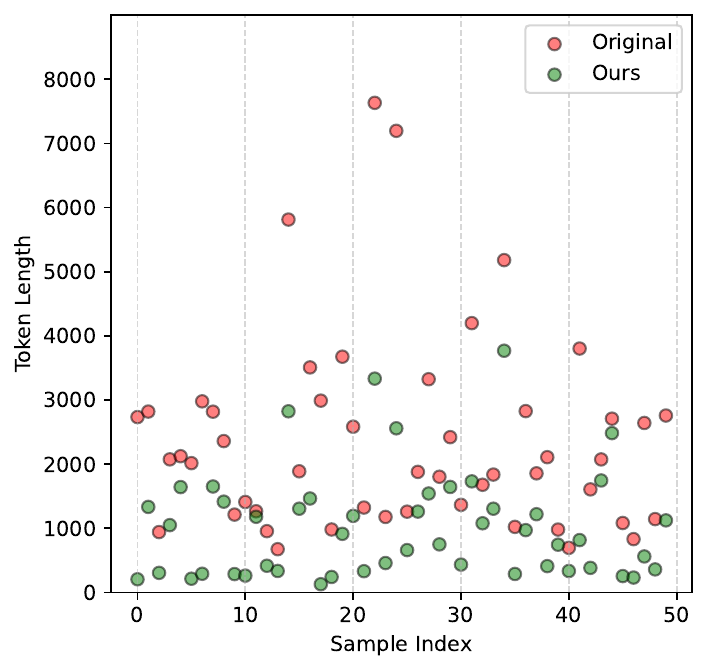}}
\caption{Visualization of the effect on token length. We can see TH2T significantly reduces response length.}
\label{fig:case_samples_len_scatter}
\end{figure*}
\section{More Experimental Results}
\subsection{Results on Llama Model}
We provide detailed experimental results of the R1-Distilled-Llama-8B model evaluated on 4 mathematical benchmarks (i.e., GSM8K, MATH, AIME2024, and OmniMath) and 1 QA benchmark (i.e., GPQA). We report accuracy and average token length for comparison in Tab.~\ref{tab:main-llama}.

\subsection{Ablation on Training Steps.} 
We examine the dynamics of accuracy and average token length during the training process (in Fig.~\ref{fig:train_process}), and obtain two interesting findings: 1) As training steps increase, accuracy initially declines, then returns to the original level, with occasional positive gains during the process. 2) For hard tasks like MATH, the response length exhibits a reverse increase after exceeding a certain number of steps. This inspires us to set the training steps to 6400 to prevent model degradation.

\subsection{Visualization of the Effect on Token Length}
We collect the first 50 samples from GSM8K and MATH test set in sequential order, and compare the output token length distribution between our TH2T and the original DeepSeek-R1-Distill-Qwen-7B in Fig.~\ref{fig:case_samples_len_scatter}. We can see TH2T significantly reduces response length.

\section{More Experimental Details}
\subsection{Dataset Construction}
The datasets used in this paper incorporate difficulty level labeling as a key consideration, so we mainly use mathematical reasoning tasks. Our TH2T is a two-stage strategy, with tailored dataset construction for each phase, followed by sequential fine-tuning. 1) During the first stage, we use the GSM8K and MATH training datasets to obtain $\mathcal{D}_{0}$ and $\mathcal{D}_{1}$, respectively. The former is generally considered simple, whereas the latter is regarded as difficult and is explicitly categorized into difficulty levels ranging from 1 to 5, and we pick samples between lv.3 and lv.5. After filtering out errors, $\mathcal{D}_{0}$ and $\mathcal{D}_{1}$ are balanced at a 1:1 ratio. After injecting difficulty hypnosis $H$, we construct $\mathcal{D}$ of 6.4K samples. 2) In Stage-2, $\mathcal{D}_{1}$ are injected with redundancy hypnosis $\smash{\bar{H}_1}$ and looping hypnosis $\smash{\bar{H}_2}$, forming a final training dataset. We observe that the meticulous composition ratio of the constructed dataset does not significantly affect the final performance. Hence, we combine $\mathcal{D}_{0}$ and $\mathcal{D}_{1}$ in a ratio of 1:1. Meanwhile, the redundancy hypnosis and looping hypnosis are also injected into $\mathcal{D}_{1}$ in a ratio of 1:1. The overall composition is set as 2:1:1, yielding $8k$ samples in total.

\subsection{Benchmarks}
We evaluate on 4 mathematical benchmarks and 1 QA benchmark. The main results are on the test set of in-distribution GSM8K and MATH. Meanwhile, AIME2024 and Omni-MATH are employed as out-of-distribution test sets. For Omni-MATH, we select 30 samples from difficulty ranges 1-2 and 4-5 as easy/hard instances, while AIME2024 is considered hard. To explore generalization ability in other domains, we include the GPQA benchmark to test performance on QA tasks.

\subsection{Metrics}
Main results are assessed on accuracy and the average number of output tokens. To substantiate the research objectives, we incorporate the success rate of the difficulty awareness. We also report the redundancy cognition optimization, which includes the reduction of redundant structures (e.g., ``Wait'', ``Alternatively'', or ``Check''), as well as the reflection looping ratio. To examine whether our method instills a genuine sense of difficulty cognition, we investigate the model's internal state, i.e., the confidence value of the first generated token.

\subsection{Implementation Details}
\label{sec:implementation_details}
We adopt LoRA, a widespread LLM fine-tuning approach, to train our models with a rank $r=8$, scaling parameter $\alpha=16$, and learning rate $r=1\times10^{-5}$. All experiments are conducted on 4$\times$NVIDIA H800 GPU (80GB). For the decoding strategy, we employ greedy decoding. The maximum generation length is set to 8192 for GSM8K and MATH500, and 16,384 for AIME2024, OmniMath, and GPQA. The random seed is fixed to 42.

\section{More Explanation on Motivation}
\label{sec:more_motivation}
\begin{figure*}[t!]
    \centering
    \subfloat[Overlap of correct responses on GSM8K]{\includegraphics[width = 0.3\textwidth]{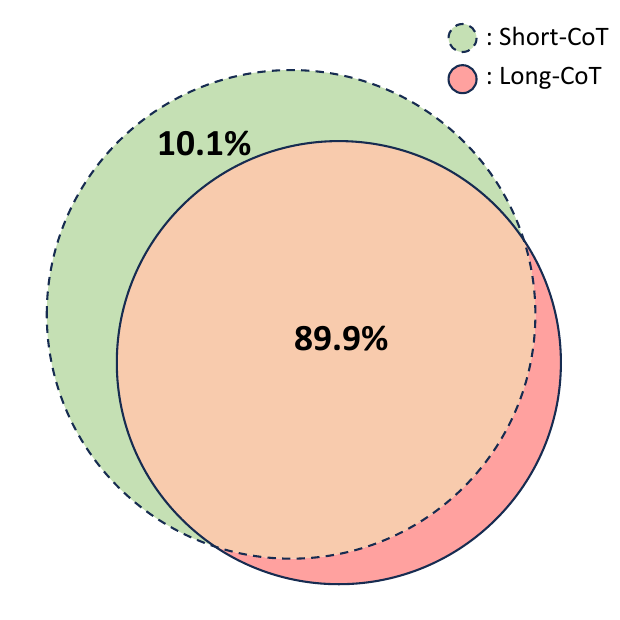}}
    \subfloat[Accuracy gain and token length scaling]{\includegraphics[width = 0.4\textwidth]{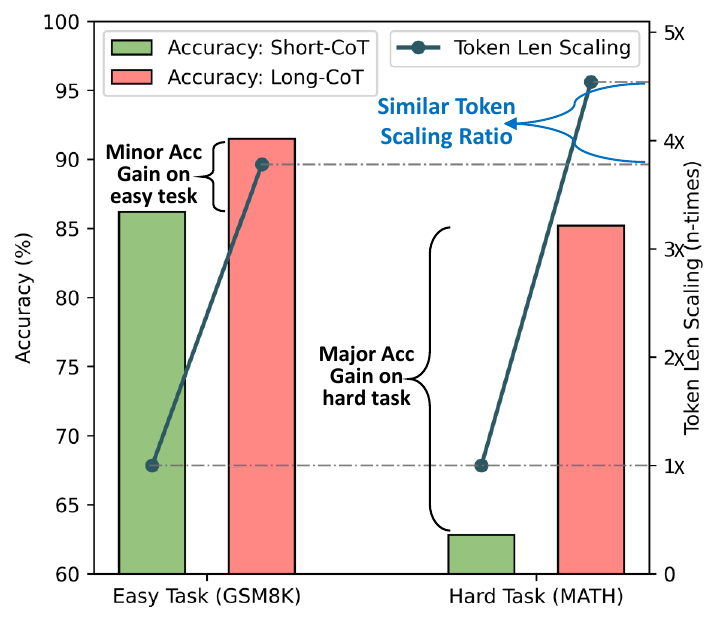}}
    \caption{Short-CoT model (Qwen2.5-7B) \textit{vs} LRM (DeepSeek-R1-Distill-Qwen2.5-7B). Left: Short-CoT replicates up to 89.9\% correct responses of Long-CoT on the easy task. Right: under a similar high token length scaling ratio (around 400\%), the easy task only witnesses a minor accuracy gain compared with the hard task. This comparison raises a reasonable speculation that LRMs fall short of perceiving task complexity, since massive inference resources on easy ones are superfluous and lead to limited performance gain.}
    \label{fig:ShortCoT_vs_LongCoT}
\end{figure*}

We investigate the efficiency issue from a suspicion: \textit{ Are current LRMs treating various task types equally with identical response strategy?} We compare the accuracy gain and inference length scaling ratio between Short-CoT model (Qwen2.5-7B) and Long-CoT model (DeepSeek-R1-Distill-Qwen2.5-7B) on easy GSM8K and hard MATH-500 benchmarks. From Fig.~\ref{fig:ShortCoT_vs_LongCoT}, we obtain \textbf{three findings}:
\begin{itemize}
\item Across the easy and difficult tasks, the current reasoning model shows similar token length scaling ratio (i.e., both around 400\%), which verifies our initial suspicion that LRMs are allocating similar additional computing resources to tasks with varying difficulty levels.
\item Despite incurring minimal costs, basic LLM can still replicate the majority of correct responses (89.9\%) of LRM's on the easy tasks, which is why we claimed that $\theta_S$ can replicate the majority correct responses of $\theta_L$ on $\mathcal{Q}_0$ in Sec.\ref{sec:stage_1}. 
\item Compared with difficult tasks, the easy task only witnesses a minor accuracy gain (5.3\% \textit{vs} 22.4\%) under the same computation resource allocation strategy, which is not what we expected. We aim to maintain this marginal performance improvement while keeping the associated token length increase at a minimal level.
\end{itemize}

\noindent These raise a reasonable speculation: \textbf{LRMs fall short of perceiving task complexity, since massive inference resources on easy ones are superfluous and lead to limited performance gain}. As discussed in Results on Redundancy Cognition of the main paper, we observe severe redundant structures in LRM's responses to hard tasks, including superfluous reflection along with a remarkably high proportion of reasoning loops in incorrect answers. These findings lead to the \textbf{two objectives} of this paper: 
\begin{itemize}
\item Enabling LRMs to think how to think, i.e., autonomously recognizing and adapting to variant levels of problem difficulty (results and analysis are presented in the Results of Difficulty Cognition section of the main paper). 
\item Mitigating LRM overthinking by promoting redundancy cognition of internal reasoning structure (results and analysis are presented in the Results of Redundancy Cognition section of the main paper).
\end{itemize}

\section{Prompts}
\label{sec:prompts}
For reproducibility, we provide the complete list of prompts adopted in our experiments.

\subsection{Difficulty level evaluation}
\begin{figure*}[h]
    \centering
    \includegraphics[width = 0.75\textwidth]{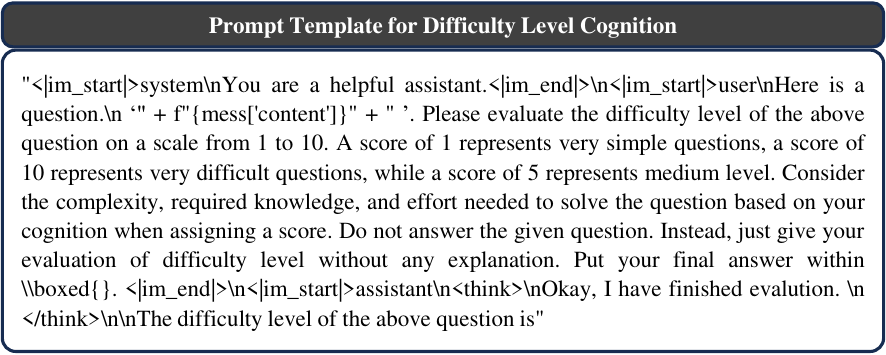}
    \caption{Prompt Template for Difficulty Level Evaluation}
    \label{fig:prompt_difficulty_evaluation}
\end{figure*}

Fig.~\ref{fig:prompt_difficulty_evaluation} shows the prompt template we used during difficulty level evaluation.

\subsection{Baseline}
\begin{figure*}[h]
    \centering
    \includegraphics[width = 0.75\textwidth]{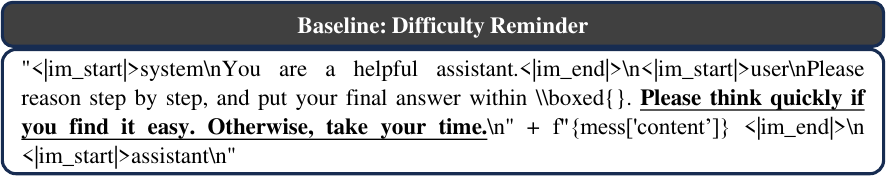}
    \caption{Setting of baseline -- Difficulty Reminder}
    \label{fig:prompt_difficulty_reminder}
\end{figure*}

\begin{figure*}[h]
    \centering
    \includegraphics[width = 0.75\textwidth]{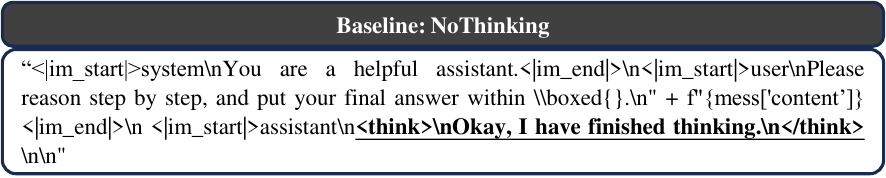}
    \caption{Setting of baseline -- NoThinking}
    \label{fig:prompt_NoThinking}
\end{figure*}

Fig.~\ref{fig:prompt_difficulty_reminder} is the difficulty reminder for the baseline \textbf{D-Prompt}, which is also adopted in the pre-experiment. Fig.~\ref{fig:prompt_NoThinking} is the setting of baseline \textbf{NoThinking}.


\section{Design of Determinator}
\label{sec:determinator}
\begin{figure*}[h]
    \centering
    \includegraphics[width = 0.75\textwidth]{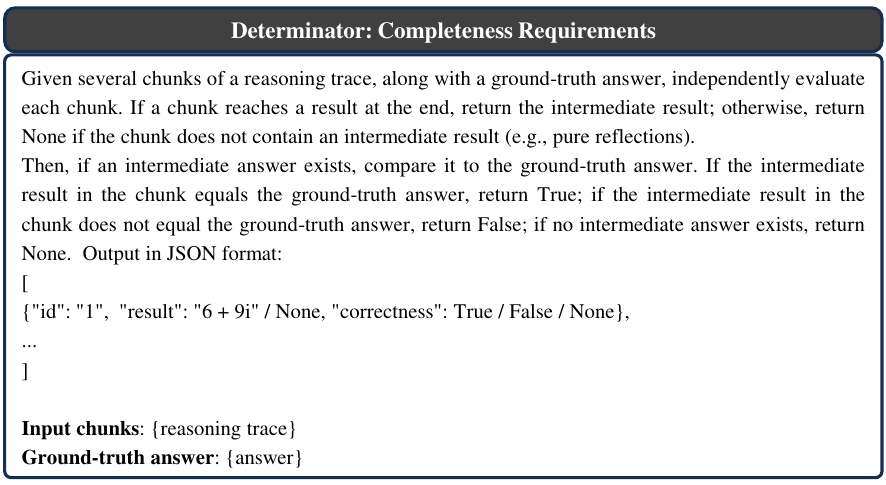}
    \caption{Design of Determinator -- Completeness Requirements.}
    \label{fig:prompt_determinator_completeness}
\end{figure*}

\begin{figure*}[h]
    \centering
    \includegraphics[width = 0.8\textwidth]{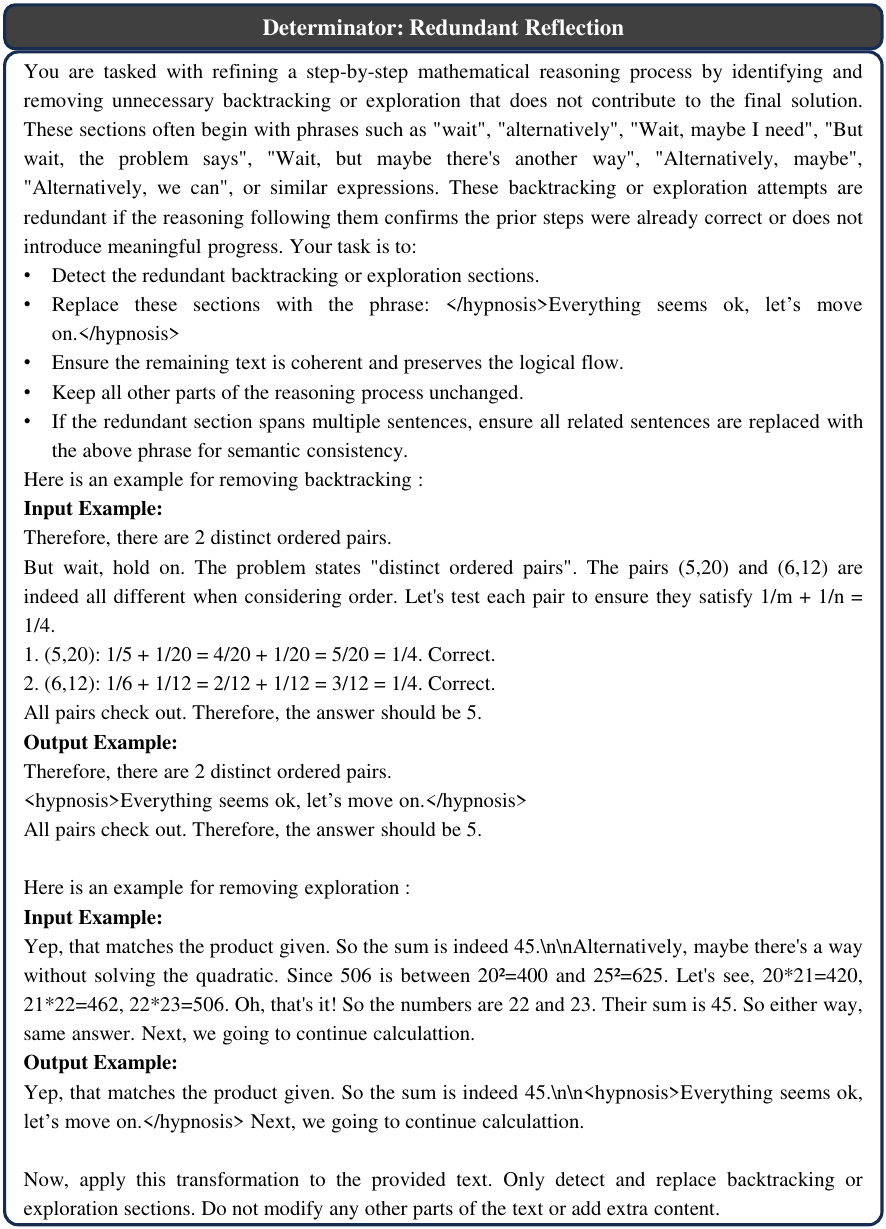}
    \caption{Design of Determinator -- Redundant Reflection.}
    \label{fig:prompt_determinator_reflection}
\end{figure*}

\begin{figure*}[h]
    \centering
    \includegraphics[width = 0.8\textwidth]{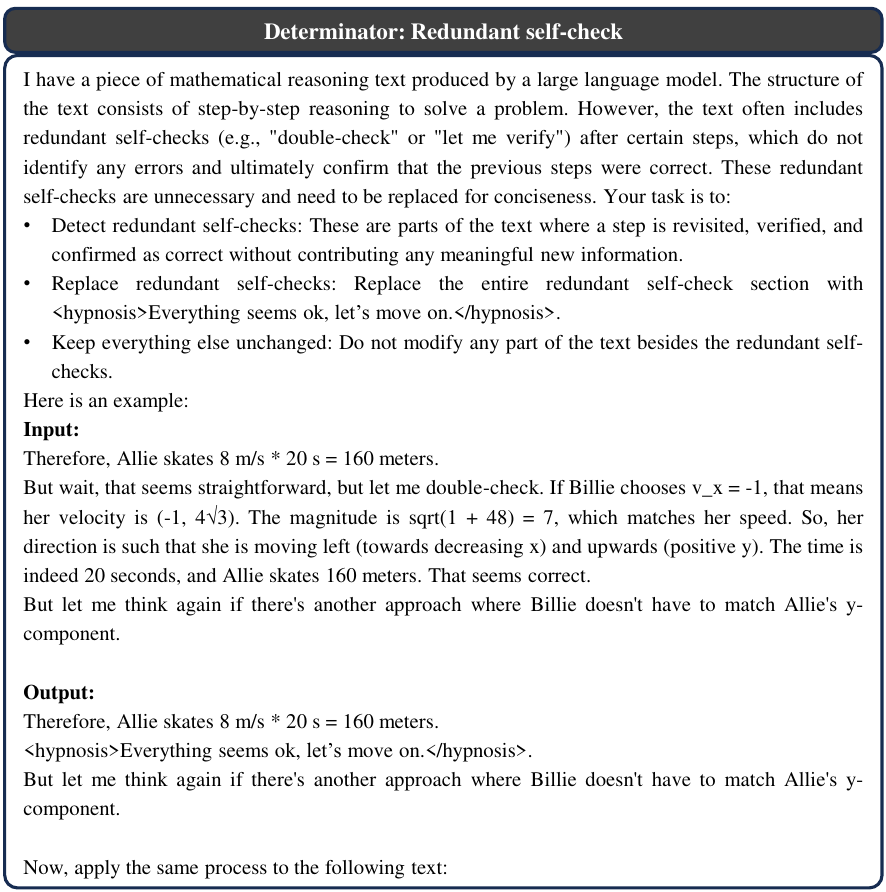}
    \caption{Design of Determinator -- Redundant self-Check.}
    \label{fig:prompt_determinator_check}
\end{figure*}

During Stage 2 of TH2T, we design a Determinator. To identify redundancy point, i.e., $\textit{Determinator}(T_{m}) \equiv T_{i}^{'}$, the contribution effectiveness of reflective structures is assessed as shown in Fig.~\ref{fig:prompt_determinator_reflection} and Fig.~\ref{fig:prompt_determinator_check}. Furthermore, when injecting hypnosis, the completeness requirement in Fig.~\ref{fig:prompt_determinator_completeness} is adopted.

\section{Case Study}
\begin{figure*}[h]
    \centering
    \includegraphics[width = 0.9\textwidth]{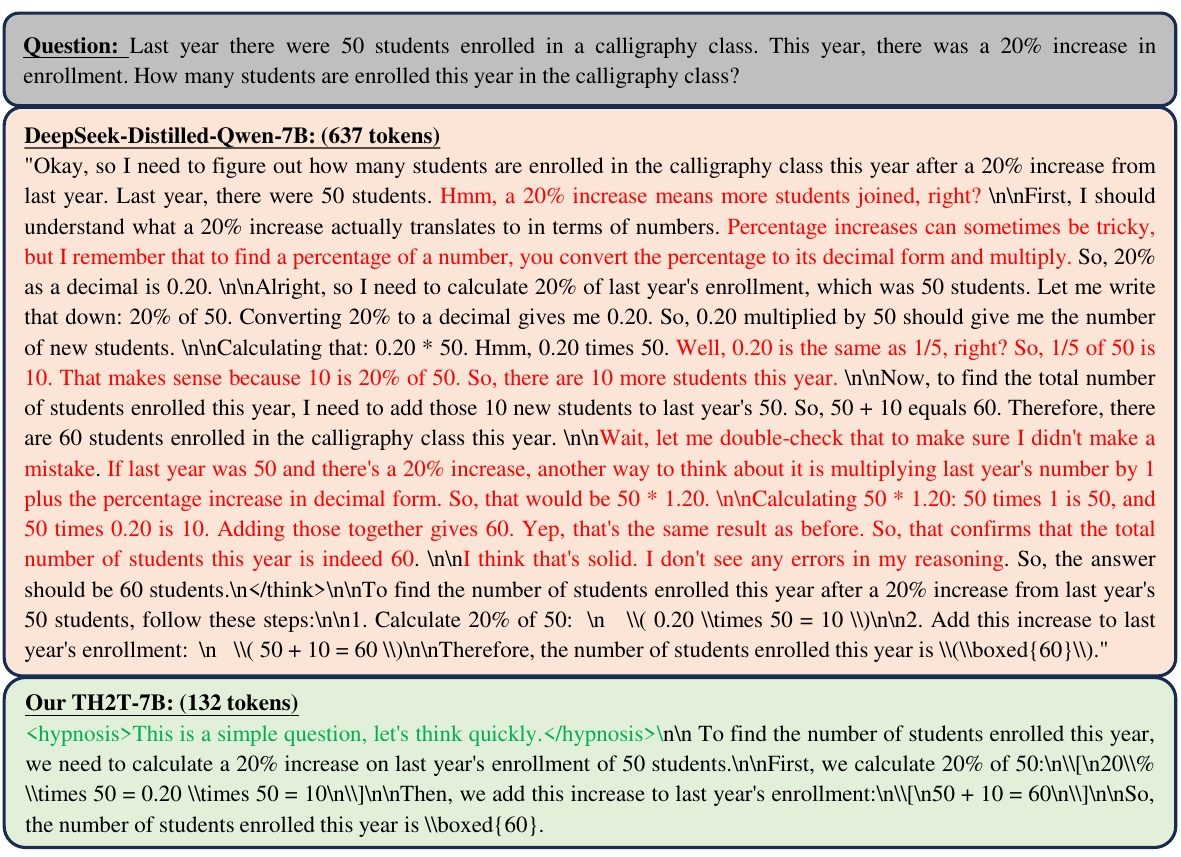}
    \caption{An example demonstrating TH2T’s native difficulty cognition capability and concise response, where an easy question from GSM8K is recognized as \textit{"easy"} with a length drop from 637 tokens to 132 tokens.}
    \label{fig:case_study_easy_easy}
\end{figure*}

\begin{figure*}[h]
    \centering
    \includegraphics[width = 0.9\textwidth]{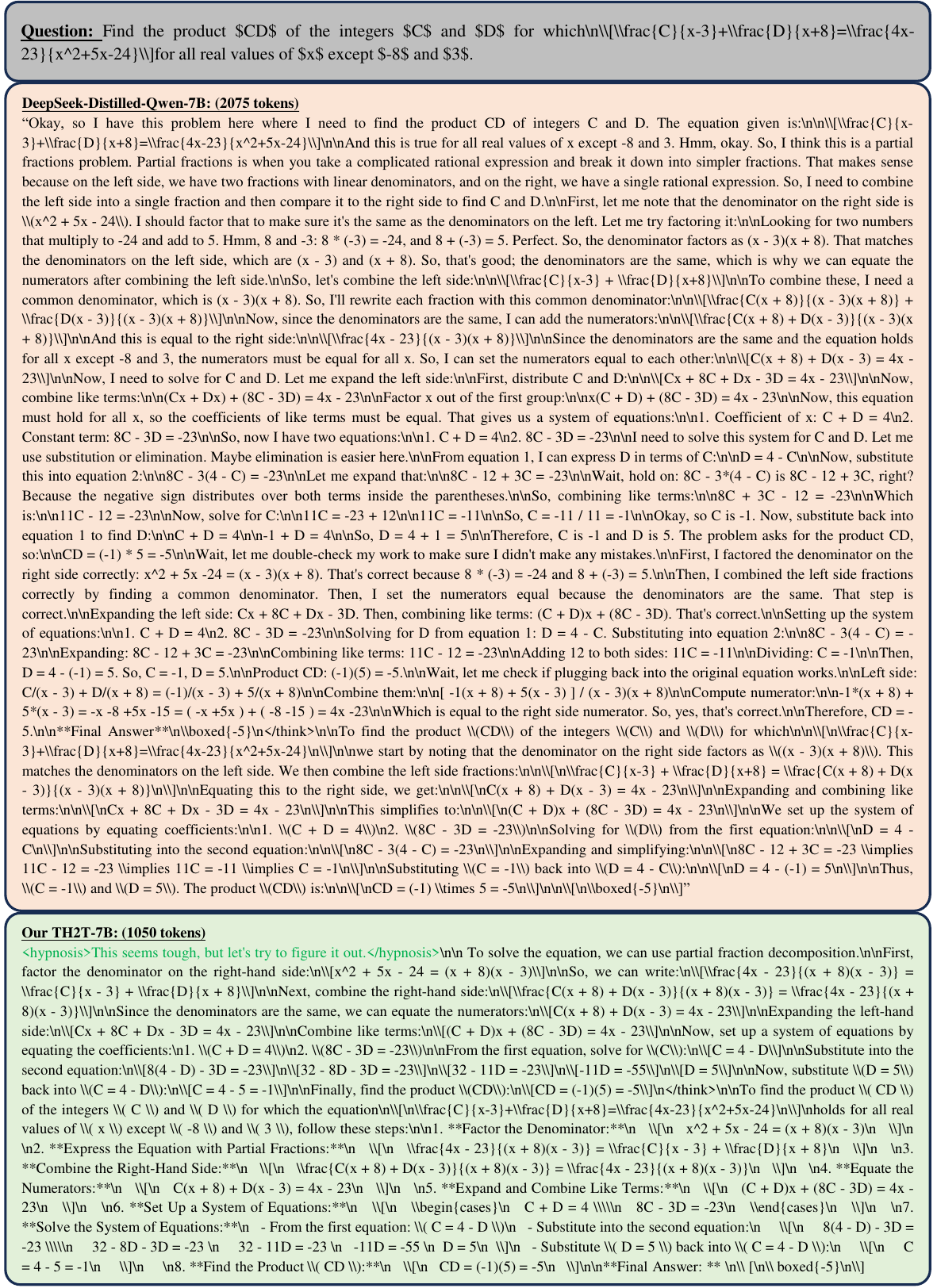}
    \caption{An example demonstrating TH2T’s native difficulty cognition capability and concise response, where an hard question from MATH is recognized as \textit{"tough"} with a length drop from 2075 tokens to 1050 tokens. We can see that the generated difficulty hypnosis in the response is not always the same as the setting in data construction, indicating generalization.}
    \label{fig:case_study_hard_hard}
\end{figure*}

\begin{figure*}[h]
    \centering
    \includegraphics[width = 0.9\textwidth]{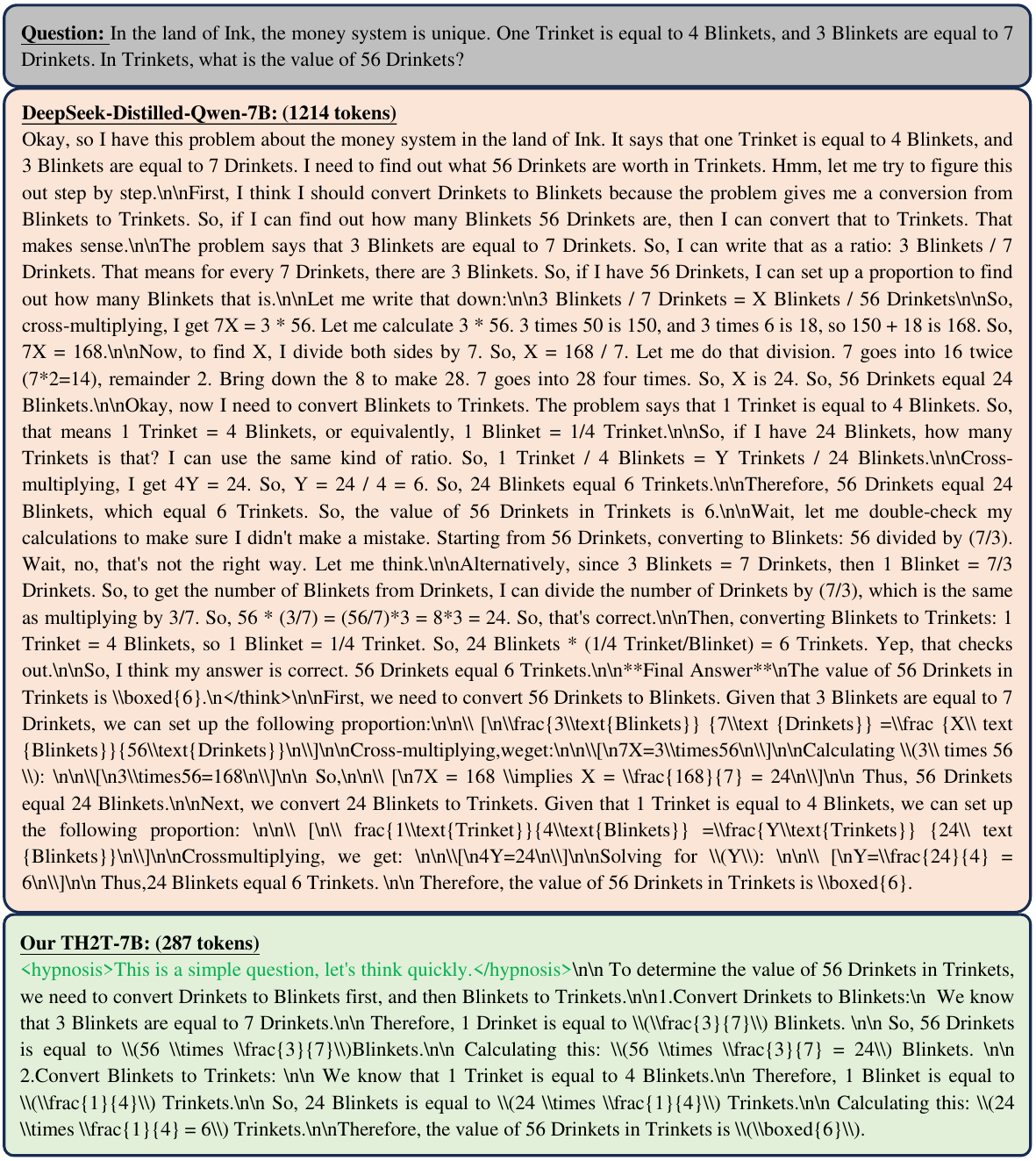}
    \caption{An example demonstrating TH2T’s native difficulty cognition capability and concise response, where a hard question from MATH is recognized as \textit{"simple"} with a length drop from 1214 tokens to 287 tokens. We can see some difficulty calibration phenomenon, further reducing the response length of difficult questions.}
    \label{fig:case_study_hard_easy}
\end{figure*}

Three concrete cases are provide in Fig.~\ref{fig:case_study_easy_easy}, Fig.~\ref{fig:case_study_hard_hard} and ~\ref{fig:case_study_hard_easy}.

\end{document}